\documentclass[letterpaper,10pt,conference]{ieeeconf}

\IEEEoverridecommandlockouts
\usepackage{amsmath,amssymb}
\usepackage{graphicx}
\usepackage{adjustbox}
\usepackage{booktabs}
\usepackage{multirow}
\usepackage{wrapfig}
\usepackage{url}
\usepackage{xcolor}
\usepackage{microtype}
\usepackage[utf8]{inputenc}
\usepackage{algorithm}
\usepackage{tabularx} 
\usepackage{algorithmic}

\newtheorem{theorem}{Theorem}
\newtheorem{definition}{Definition}
\usepackage[font=small,labelfont=bf]{caption}
\usepackage[dvipsnames]{xcolor}

\definecolor{ocre}{RGB}{243,102,25}
\newcommand{\husseinold}[1]{{\color{black} #1}}
\newcommand{\ihabold}[1]{{\color{black} #1}}

\definecolor{mypurple}{RGB}{176,175,243} 

\title{
Learning Conservative Neural Control Barrier Functions from Offline Data
}

\author{Ihab Tabbara$^{1}$ and Hussein Sibai$^{2}$
\thanks{$^{1}$Ihab Tabbara is a PhD student at Washington University in St. Louis, St. Louis, MO 63130, USA.
        {\tt\small i.k.tabbara@wustl.edu}}%
\thanks{$^{2}$Hussein Sibai is an Assistant Professor in the Department of Computer Science at Washington University in St. Louis, St. Louis, MO 63130, USA.
        {\tt\small sibai@wustl.edu}}%
}

\begin{document}

\maketitle
\begin{abstract}
Safety filters, particularly those based on control barrier functions, have gained increased interest as effective tools for safe control of dynamical systems. Existing correct-by-construction synthesis algorithms for such filters, however, suffer from the curse-of-dimensionality. Deep learning approaches have been proposed in recent years to address this challenge.  
In this paper, we add to this set of approaches 
an algorithm for training neural control barrier functions from offline datasets. Such functions can be used to design constraints for quadratic programs that are then used as safety filters. Our algorithm trains    
these functions so that the system is not only prevented from reaching unsafe states, but is also disincentivized from reaching out-of-distribution ones, at which they would be less reliable. It is inspired by Conservative Q-learning, an offline reinforcement learning algorithm. 
We call its outputs Conservative Control Barrier Functions (CCBFs). 
Our empirical results demonstrate that CCBFs outperform existing methods in maintaining safety
while minimally affecting task performance. 
Source code is available at https://github.com/tabz23/CCBF.



\end{abstract}

\section{Introduction}

Ensuring that control systems satisfy safety specifications is critical in various applications, 
such as autonomous driving~\cite{autonomous_driving_a_crash_explained_in_detail_2019} and robotic surgery~\cite{robotic_surgery_survey_2019}. Safety filters 
have emerged as promising tools for enforcing safety constraints~\cite{hsu2023safety}. 
Control barrier functions (CBF) allow the design of efficient safety filters in the form of quadratic programs (QP) that preserve the forward invariance of sets \husseinold{of states disjoint from a user-defined {\em failure} set of states, such as collisions,}  while minimally altering (potentially unsafe) task-achieving nominal control~\cite{cbf_overview_2018,ames2016control}. 
However, synthesizing CBFs using correct-by-construction techniques, such as sum-of-squares programming~\cite{ahmadi_majumdar_applications_2016,clark2021verification}, do not scale beyond few dimensions. 

To address this challenge, researchers have proposed  data-driven methods for learning safety filters~\cite{learned_certificates_survey_chuchu_2023}, which have been applied in settings with  
uncertain dynamics~\cite{dawson2022safe,lindemann2024learning} and high-dimensional observations,  
such as images~\cite{abdi2023safe,xiao2022differentiable,nakamura2025generalizing} and point clouds~\cite{he2024agile,keyumarsi2023lidar}. However, these filters lack the theoretical guarantees that their formally verified or synthesized counterparts have. Moreover, when deployed, by design, they will alter the nominal controls resulting in a shift between the distribution of trajectories used for their training and the ones visited during deployment. It is often assumed that the filter is trained iteratively until convergence: in each iteration, it is used to generate new trajectories whose states get labeled as safe\husseinold{, i.e., belong to a forward invariant set disjoint from the failure set,} or unsafe\husseinold{, i.e., belong to the complement of that forward invariant set}, and then the filter gets retrained accordingly~\cite{how_to_train_cbfs_2024}. After convergence, the distribution of states used to train the filter would match the distribution of states it will encounter during deployment. Consequently, one can use classic generalization bounds to obtain  probabilistic guarantees on the correctness of the learned filter. However, in various safety-critical settings, it can be challenging and expensive to collect new trajectories.
This highlights the need for learning safety filters from offline data, while addressing or mitigating the resulting distribution shift between training and deployment.

\begin{figure}[ht]
\centering
\includegraphics[width=\columnwidth]{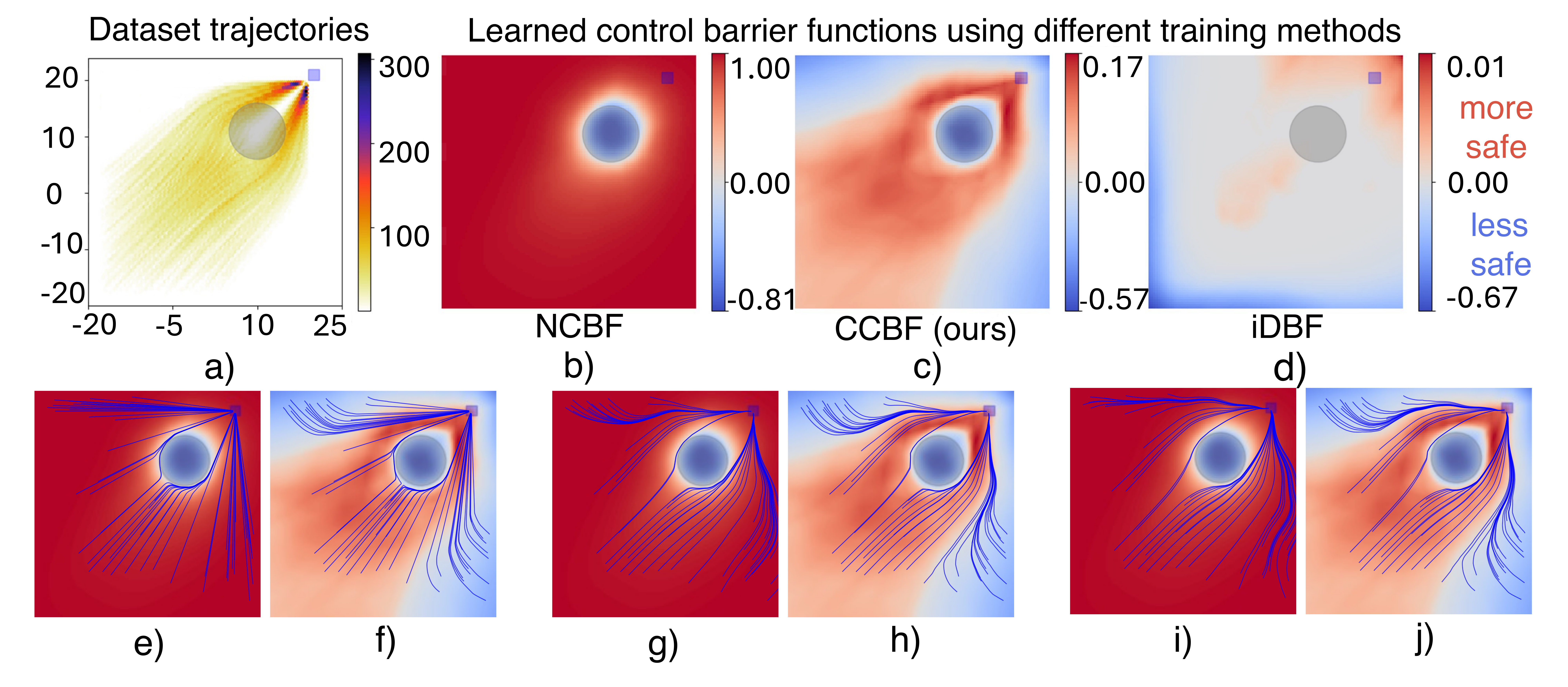}
\caption{  \textbf{Learned barrier functions and trajectory rollouts.} \\
(a) Offline dataset trajectories with an obstacle (\textcolor{gray}{gray}) and a goal (\textcolor{mypurple}{purple})
(b–d) 
NCBF, CCBF (ours), and iDBF, respectively. 
(e–j) Rollouts (blue) using three CBF-QP policies, with nominal controllers ordered left to right as PD, BC, and BC-Safe (defined in Section~\ref{sec:nominal_controllers}). 
For each pair of figures, the left panel shows the results of using NCBF as the CBF in CBF-QP policy and the right panel shows the results of using CCBF instead. 
CCBF consistently balances safety with goal-reaching by accounting for both the obstacle and the training distribution, 
whereas NCBF emphasizes only obstacle avoidance and iDBF is very conservative.
  }
\label{fig:NCBF-CCBF-iDBF-trajectories-visualization}
\end{figure}

Offline reinforcement learning (RL) has been studied extensively in the past few years~\cite{offline_RL_tutorial_levine_2020}. Several algorithms have been proposed to learn optimal policies from offline datasets of trajectories in that literature~\cite{kumar2020conservative, Behavior_regularized_offline_RL, Uncertainty_weighted_actor_critic_2021, kumar2019stabilizing, lee2022coptidice}. The goal is to obtain policies that outperform the ones used to generate the data or those which can be obtained using behavior cloning (BC), while accounting for distribution shift. 
In this paper, we aim to train safety filters using  offline datasets of safe and unsafe trajectories that alter unsafe nominal controls to 
safe ones.
For that purpose, we introduce Conservative Control Barrier Functions (CCBFs), a method for training neural CBFs that is inspired by conservative Q-learning (CQL)~\cite{kumar2020conservative}, a prominent offline RL algorithm. The core insight is that the learned neural CBF should avoid over-estimating the safety of states and actions not seen in the dataset. 
\ihabold{ Our proposed method exhibits this property as can be seen in Figure \ref{fig:NCBF-CCBF-iDBF-trajectories-visualization}}. This objective is similar to the CQL objective, which minimizes the $Q$-values for randomly sampled actions at the dataset states in order to learn a $Q$-function that lower bounds the true one. That addresses the common problem of overestimating the $Q$-values for unseen state-action pairs in offline RL. 
By analogously minimizing the barrier values of states reached by random actions starting from the dataset states, 
CCBFs avoid overestimating the safety of unseen states. 
We evaluate our approach against three baseline methods in four environments with unknown dynamics and with varying state dimensions. Our results show that CCBFs outperform the baselines in achieving safety without sacrificing task performance  while being easy to train and not requiring significant tuning of the hyperparameters. 

\section{Related Work}
\label{sec:related_work}
\paragraph{Learning safety filters from expert demonstrations}
Several algorithms have been proposed recently to train neural CBFs from offline datasets~\cite{lindemann2024learning,yu2025estimating,so2024train}. They differ in their objectives, their dataset generation process, their assumptions, and their guarantees, if any. 
\cite{castaneda2023distribution} and \cite{kang2022lyapunov} proposed learning safety filters to prevent control systems from reaching OOD states.  \cite{castaneda2023distribution} proposed an algorithm for training neural CBFs, termed in-distribution barrier functions (iDBFs), using an offline dataset of safe trajectories to prevent the control system from reaching OOD states. To train an iDBF, the algorithm first trains a BC policy using the same dataset. 
Then, for each state in the training dataset, the algorithm randomly samples a set of actions, and if the BC policy assigns a sampled action a probability density below a user‑defined threshold, 
the next state reached by that action is labeled OOD and appended to the dataset. It finally uses an existing approach for training neural CBFs (e.g.,~\cite{dawson2022safe}) to train the iDBF using the augmented dataset, treating the InD states as safe and the OOD states as unsafe. \cite{kang2022lyapunov} proposed Lyapunov Density Models, combining control Lyapunov functions \cite{sontag1999control} and density models \cite{ostrovski2017count}, which are also learned from data to keep a control system from reaching OOD states.  More recently, \cite{Estimating_Control_Barriers_from_Offline_Data_2025} proposed an algorithm for learning neural CBFs from an offline dataset consisting of  a mix of labeled and unlabeled trajectories by leveraging cost-sensitive classification \cite{cost_sensitive_classification_2020_pmlr} to assign safety labels to the unlabeled demonstrations. 

Several other works proposed algorithms for synthesizing CBFs from expert safe demonstrations, providing formal correctness guarantees through Lipschitz continuity and coverage-based proofs~\cite{robey2020learning,lindemann2024learning}.


In this paper, we aim to train neural CBFs from offline datasets of both safe and unsafe demonstrations that can be used as safety filters for control systems to avoid unsafe states and disincentivize reaching OOD ones while minimally affecting task progress. 

\paragraph{Safe offline RL}

Safe offline RL is a branch of offline RL that aims to train policies that satisfy safety constraints while maximizing reward using offline datasets. It tackles a similar problem to ours: given a set of trajectories of a Constrained Markov Decision Process (CMDP), design an optimal and safe policy that avoids distribution shift~\cite{offline_rl_review_levine_2022}. However, it considers soft safety constraints which require the expected cost of the trajectories to remain below a user-defined threshold, in contrast with  the hard constraints of avoiding failure states that we consider. Several approaches incorporated safety filters into the safe offline RL pipeline \cite{zheng2024safe,zhao2025reachability}.
A close work to ours is the recently proposed method FISOR \cite{zheng2024safe}, which similarly employs a hard state-wise constraint formulation for safe offline RL. It first adapts Hamilton-Jacobi reachability to learn a value function that defines the backward reachable set (BRS) of a known constraint set (aka  failure set) from the offline dataset. It uses Implicit Q-learning~\cite{implicit_q_learning_kostrikov_levine_ICLR_2022} to prevent over-optimistic safety value estimation that can result from querying OOD actions in offline RL settings. It then trains a weighted BC policy that maximizes rewards within the feasible region in the state space, i.e., the complement of the backward reachable set.
We instead take a different approach and assume that the states in the dataset are labeled as safe (belong to a controlled forward invariant set disjoint from the failure set) or unsafe (i.e., not safe), train a neural CBF with an additional loss term inspired from CQL~\cite{kumar2020conservative} to avoid over-optimistic CBF values for OOD states, and test its performance as a safety filter for various nominal controllers that are learned separately from the same dataset using various safe offline RL algorithms. 

\section{Preliminaries}

In this paper, we consider nonlinear control-affine  systems. 
\begin{definition}[Control-affine systems]
    A nonlinear control-affine system can be modeled  using an ordinary differential equation (ODE) of the form:
    \begin{equation}\label{eq:system}
        \dot{x} = f(x) + g(x)u,
    \end{equation}
    where \( x(t) \in \mathcal{X} \subseteq \mathbb{R}^n \) represents the state, and \( u(t) \in \mathcal{U} \subseteq \mathbb{R}^m \) denotes the control input at time \( t \). The functions \( f: \mathcal{X} \to \mathbb{R}^n \) and \( g: \mathcal{X} \to \mathbb{R}^{n \times m} \) are assumed to be locally Lipschitz continuous.
\end{definition}

A set of states $S \subseteq \mathcal{X}$ is called {\em controlled forward invariant} with respect to system~(\ref{eq:system}) if there exists a control policy that makes it  {\em forward invariant}, i.e., all trajectories starting in $S$ remain inside it.

We assume that there is a user-defined set of failure states, such as collisions or violations of speed limits of autonomous vehicles, that a safe controller should prevent the system from reaching. Given a controlled forward invariant set $S$ that is disjoint from the failure set, we say that the states in $S$ are {\em safe} (and $S$ is the safe set) and the states in $\bar{S} := \mathcal{X}$\textbackslash $S$ are {\em unsafe} (and $\bar{S}$ is the unsafe set). Note that $\bar{S}$ contains the failure set and might also contain additional states, particularly when there are input constraints or the system has a high relative degree~\cite{how_to_train_cbfs_2024,hamiltonjacobi}. Thus, safety is defined as the invariance to a set disjoint from the failure set.  The largest safe set $S$ can be obtained using Hamilton-Jacobi (HJ)  reachability analysis~\cite{hamiltonjacobi}. Unfortunately, existing algorithms for HJ reachability do not generally scale beyond few dimensions~\cite{level_set_toolbox_flexible_2008}. 

%

\subsection{Control barrier functions}

\begin{definition}[Control barrier functions~\cite{cbf_overview_2018}]
    A function \( B: \mathcal{X} \to \mathbb{R} \) is a \emph{control barrier function} for system \eqref{eq:system} if it is continuously differentiable and satisfies: $\forall x \in \mathcal{D} \subseteq \mathcal{X}$, 
    \begin{align}
        \exists u \in \mathcal{U} \text{ such that } \dot{B}(x,u) + \alpha (B(x)) \geq 0, \label{eq:cbf-def-cond-ascent}
    \end{align}
   where the super-level set \( B_{\geq 0} := \{x \ |\ B(x) \geq 0\} \) belongs to \(\mathcal{D} \), and \( \alpha: \mathbb{R} \to \mathbb{R} \) is a locally Lipschitz extended class-\( \mathcal{K}_{\infty} \) function, i.e., it is strictly increasing and satisfies \( \alpha(0) = 0 \).
\end{definition}

A CBF $B$ for system~(\ref{eq:system}) specifies the controls  that guarantee the forward invariance of $B_{\geq 0}$,  
as stated in the following theorem. If the set of unsafe states is disjoint from $B_{\geq 0}$ and the system starts from $B_{\geq 0}$, then the system can be kept safe by following such controls. 




\ihabold{
\begin{theorem}[\cite{cbf_overview_2018}]
Any Lipschitz continuous control policy $\pi: \mathcal{D} \to \mathcal{U}$ satisfying  
$\forall x \in \mathcal{D}$, $\pi(x) \in K_{\text{cbf}} := \{u \in \mathcal{U} \mid \nabla B(x)(f(x) + g(x)u) 
+ \alpha(B(x)) \geq 0\}$
makes the set $B_{\geq 0}$ forward invariant.  
\end{theorem}
}
Given a reference controller \( \pi_{\text{ref}}: \mathcal{X} \to \mathcal{U} \) that does not necessarily satisfy safety constraints, a safety filter using the CBF \( B \) can be employed to modify its unsafe control choices. Specifically, a quadratic program (QP) can be formulated to determine the closest safe control to \( \pi_{\text{ref}}(x) \) \cite{ames2016control}, resulting in what is denoted by a CBF-QP policy $\pi_{\text{safe}}$, as shown below:  
\ihabold{
\begin{equation}\label{eq:qp}
    \begin{aligned}
    \pi_{\text{safe}}(x) &:= \arg\min_{u \in \mathcal{U}} \lVert u - \pi_{\text{ref}}(x) \rVert^2 \\
    &\quad \text{s.t. } \nabla B(x)(f(x) + g(x)u) + \alpha(B(x)) \geq 0. 
    \end{aligned}
\end{equation}}

Under some non-restrictive conditions, it can be proven that $\pi_{\text{safe}}$ is locally Lipschitz continuous for $x \in \text{Int}(B_{\geq 0})$, the interior of the set $B_{\geq 0}$~\cite{ames2016control}.

\subsection{Offline reinforcement learning}
\label{sec:offline_RL_prelims}

The goal of offline RL is to design optimal control policies
 from offline datasets without further interaction with the environment. 
 Conservative Q learning (CQL) \cite{kumar2020conservative} is a prominent offline RL approach with several variants. The first variant, which is the one we are interested in here and simply call CQL for the rest of the paper, learns a $Q$ function from the offline dataset corresponding to some policy that lower bounds its actual $Q$ function pointwise for all state-action pairs, guaranteeing not over-estimating the values of state-action pairs that are not part of the dataset. Similarly, we want to learn barrier functions which do not over-estimate the safety of states not part of the dataset.   

CQL modifies the $Q$-function learning iterative update for a policy $\pi$ by adding a term to minimize the $Q$-values of a set of state-action pairs, where the states are those in the dataset and the actions are sampled from some distribution $\mu(a|s)$, to the standard Bellman error term, as follows:
\begin{align}
\hat{Q}^{k+1} &\;\leftarrow\; \arg\min_Q \; \alpha \, \mathbb{E}_{s\sim \mathcal{D},\,a\sim\mu(a|s)}[Q(s,a)] \nonumber \\
&\quad + \tfrac{1}{2}\,\mathbb{E}_{s,a\sim \mathcal{D}}\Big[ Q(s,a) - \hat{B}^{\pi}\hat{Q}^k(s,a)\Big]^2,
\label{eq:th3.1_update}
\end{align}
where $\alpha$ is a hyperparameter, $\mathcal{D}$ is the offline dataset, 
$\hat{Q}^{k+1}$ denotes the learned $Q$-function at iteration $k+1$, $\hat{Q}^k$ is the learned $Q$-function from the previous iteration, and $\hat{B}^\pi$ is the empirical Bellman operator for policy $\pi$.
Theorem~3.1 in~\cite{kumar2020conservative} establishes that the fixed point $\hat{Q}^\pi = \lim_{k\to\infty}\hat{Q}^k$ obtained by the update in equation (\ref{eq:th3.1_update}) when the support of $\mu$ is a subset of the support of $\pi_\beta$ is a pointwise lower bound for the true $Q$-function of policy $\pi$, i.e., 
\[
\hat{Q}^\pi(s,a) \;\leq\; Q^\pi(s,a), \qquad \forall (s,a),
\]
when $\alpha$ is sufficiently large. 
Thus, the learned $Q$-function $\hat{Q}^\pi$ conservatively underestimates the true $Q$-function $Q^\pi$, preventing over-optimistic estimates for out-of-distribution actions. Previous research has already built a connection between control barrier functions and value functions~\cite{value_functions_are_CBFs,how_to_train_cbfs_2024}, so we adopt a similar analogy here 
to design a conservative loss for learning control barrier functions which do not over-estimate the safety of states unseen in the dataset.


\subsection{Problem statement}

We consider the setting where an offline trajectory dataset, that is generated using a potentially unknown and unsafe reference controller, 
is available, along with a user-specified failure set of states. 
We assume that the user has domain knowledge that allows them to label some, if not all, of the states in the dataset whether they belong to a (not exactly known) controlled forward invariant set that is disjoint from the user-specified failure set, i.e., safe, or not\footnote{One can only label the states which belong to the provided failure set as unsafe and leave the rest of the states in the dataset unlabeled. However, that may lead to a neural CBF with a small or empty super-level set, which represents the safe set. That limits the states at which the system can start and the resulting safety filter be effective.  Alternatively, we can adapt our method to training policy neural CBFs~\cite{how_to_train_cbfs_2024}, which does not require the states to be labeled as safe or unsafe to train a neural CBF with a large super-level set, but we leave that for future work.}.
Our objective is to train a neural CBF using the offline dataset that can be used to define the constraints in a quadratic program that acts as a safety filter that prevents the system from reaching the failure states. 
Consequently, 
it must also not over-estimate the safety of states unseen in the dataset which can result in trajectories reaching the failure set. 
Finally, the safety filter should not be conservative beyond what is necessary to maintain safety, i.e., as minimally altering the nominal control as possible. 

\section{Method}
Our aim is to design a CBF-QP policy from offline data. If the reference policy or the dynamics are not available, we train corresponding models using the same data. Moreover, in settings with high-dimensional observations, we train representation models and consider their latent spaces to be the state spaces based on which we train the dynamics models and the neural CBFs.
We discuss  how we train each of these components next.


\subsection{Learning latent dynamics}
\label{sec:learn_latent_dynamics}
We 
train 
a deterministic autoencoder to project high-dimensional visual observations into a low-dimensional latent space. We train it in parallel with a control-affine latent dynamics model.
The  autoencoder maps RGB images $o_t \in \mathbb{R}^{c \times h \times w}$ to latent vectors $x_t \in \mathcal{X}$ using an encoder $\mathcal{E}$, and reconstructs them using a decoder $\mathcal{G}$. We train a control-affine dynamics model of the form:
\begin{equation}
\label{eq:latent_dynamics}
x_{t+1} = x_t + (f_\theta(x_t) + g_\theta(x_t)u_t) \Delta t,
\end{equation}
where $f_\theta$ and $g_\theta$ are multilayer perceptrons parametrized by $\theta$, approximating the continuous-time dynamics model $\dot{x} = f_\theta(x) + g_\theta(x)u$.
We train the system by minimizing the following loss function using gradient descent:
$
\mathcal{L}
  := \lambda_1 \,\|o_t - \mathcal{G}\bigl(\mathcal{E}(o_t)\bigr)\|_2^{2}
  + \lambda_2 \,\|\mathcal{E}(o_{t+1}) - \hat{x}_{t+1}\|_2^{2}
  + \lambda_3 \,\|o_{t+1} - \mathcal{G}(\hat{x}_{t+1})\|_2^{2}
$, where $\hat{x}_{t+1}$ is the predicted next latent state using (\ref{eq:latent_dynamics}), and $\lambda_1$, $\lambda_2$, and $\lambda_3$ are  hyperparameters. The first term ensures that the latent vector retains sufficient information for accurate image reconstruction, the second enforces accurate latent dynamics prediction, and the third maintains consistency between predicted and ground-truth observations. During training, we optimize the autoencoder's parameters using the first loss term $\lambda_1 \,\|o_t - \mathcal{G}\bigl(\mathcal{E}(o_t)\bigr)\|_2^{2}$, while holding those parameters fixed when optimizing the dynamics model to maintain a stable latent state space.

\ihabold{Generally, we assume that the continuous time dynamics are either user-provided or that they can be approximated in the form of (\ref{eq:latent_dynamics}) accurately on the training distribution, which is the case in our experiments. Our assumption that the dynamics are control-affine is needed so that the online safety filtering problem can be formulated as a quadratic program.}

\subsection{Learning nominal controllers}
\label{sec:nominal_controllers}

We train neural reference (or nominal) controllers on the same  offline dataset we use for training the CCBF. We use behavior cloning (BC) and safe offline RL algorithms to train these controllers. We train BC-based controllers by solving the optimization problem 
$\arg\max_\beta \mathbb{E}_{(x, u) \sim \mathcal{D}} [\log \pi_\beta(u|x)]$. When we only use  the safe trajectories in the dataset in BC, we call the resulting policy BC-Safe. In the navigation environment, we also use a PD controller defined as \( \mathbf{u}_{\text{PD}} = K_p (\mathbf{x}_g - \mathbf{x}) + K_d \frac{\mathbf{u}_{\text{nom}}}{\Delta t} \), where \( \mathbf{x}_g \) is the goal position, and \( K_p, K_d \) are the proportional and derivative gains. 
For safety gymnasium environments \cite{ji2023safety}, we train BC and safe offline RL policies using DSRL \cite{dsrl}, which provides training datasets and reference implementations for safe offline RL algorithms. While such policies will be safer than the ones that result from traditional (non-safe) offline RL algorithms, they optimize for soft constraints and as we show in our experiments, generally become safer when coupled with a safety filter.
\subsection{Learning CCBFs from offline datasets}
\label{sec:learn_cbf}

We assume that a dataset of trajectories of the form \(\sigma := \{x_1, u_1, x_2, u_2, \dots, x_T, u_T\}\) is available. We define \(\mathcal{X}_{\text{safe}}\) as the set of  states labeled as safe  and \(\mathcal{X}_{\text{unsafe}}\) as the set of  states labeled as unsafe in the dataset. We denote the set of consecutive triplets \((x,u,x')\) in the dataset in which \(x' \in \mathcal{X}_{\text{safe}}\) by \(\mathcal{D}_{\text{safe}}\) and the set of triplets in which \(x \in \mathcal{X}_{\text{safe}}\) and \(x'\in\mathcal{X}_{\text{unsafe}}\) by \(\mathcal{D}_{\text{unsafe}}\).  
Given this dataset, we train a neural CCBF $B_\phi: \mathcal{X} \to \mathbb{R}$ by minimizing the following loss function:
$\mathcal{L}_{\text{CCBF}}(\phi) = \mathcal{L}_{\text{safe}}(\phi) + \mathcal{L}_{\text{unsafe}}(\phi) + \mathcal{L}_{\text{ascent}}(\phi) 
    + \mathcal{L}_{\text{descent}}(\phi) + \mathcal{L}_{\text{lip}}(\phi) + \mathcal{L}_{\text{c}}(\phi)$, 
where
$\mathcal{L}_{\text{safe}}(\phi) = \frac{w_{\text{safe}}}{|\mathcal{X}_{\text{safe}}|} \sum_{x \in \mathcal{X}_{\text{safe}}} \sigma(\varepsilon_{\text{safe}} - B_\phi(x))$,\\ 
$\mathcal{L}_{\text{unsafe}}(\phi) \!=\! \frac{w_{\text{unsafe}}}{|\mathcal{X}_{\text{unsafe}}|} \sum_{x \in \mathcal{X}_{\text{unsafe}}} \sigma(\varepsilon_{\text{unsafe}} + B_\phi(x))$, \\
$\mathcal{L}_{\text{ascent}}(\phi) \!=\! \frac{w_{\text{ascent}}}{|\mathcal{D}_{\text{safe}}|} \sum_{(x,u,x') \in \mathcal{D}_{\text{safe}}} \sigma(\varepsilon_{\text{ascent}} - \nabla B_\phi(x) (f_\theta(x) + g_\theta(x)u) - \alpha B_\phi(x))$,  \\
$\mathcal{L}_{\text{descent}}(\phi)=\frac{w_{\text{descent}}}{|\mathcal{D}_{\text{unsafe}}|}\sum_{\substack{(x,u,x') \in \mathcal{D}_{\text{unsafe}}}}\sigma(\varepsilon_{\text{descent}}+\nabla B_\phi(x)(f_\theta(x) + g_\theta(x)u) + \alpha B_\phi(x))$, \\
$\mathcal{L}_{\text{lip}}(\phi) \!=\! \frac{w_{\text{lip}}}{|\mathcal{D}|} \sum_{(x, u, x') \in \mathcal{D}} \left\|B_\phi(x') - B_\phi(x)\right\|$, \\
$\mathcal{L}_{\text{c}}(\phi) = \frac{w_{c}}{|\mathcal{X}_{\text{safe}}|} \sum_{x \in \mathcal{X}_{\text{safe}}} \tau\ln \left(\sum_{v \in V_x} \exp\left(\frac{B_\phi(x'_v)}{\tau}\right)\right)$.

Here, $\sigma$ is the ReLU activation function, and $\varepsilon_{\text{safe}}$, $\varepsilon_{\text{unsafe}}$, $\varepsilon_{\text{ascent}}$, $\varepsilon_{\text{descent}}$, $w_{\text{safe}}$, $w_{\text{unsafe}}$, $w_{\text{ascent}}$, $w_{\text{descent}}$, $w_{\text{lip}}$, $w_c$, and $\tau$ are \husseinold{non-negative real numbers chosen as} hyperparameters. The set $V_x$ is a uniformly sampled set of actions from $\mathcal{U}$ to evaluate at state $x$ as well as the action taken in the dataset at $x$. The state $x'_v$ is the state reached by the learned dynamics following action $v$ at state $x$. 
When we train without using $\mathcal{L}_{\text{c}}$, we call the result a neural control barrier function (NCBF).

The loss terms $\mathcal{L}_{\text{safe}}$ and $\mathcal{L}_{\text{unsafe}}$   drive the CCBF to be positive at safe states and negative at unsafe states. The terms $\mathcal{L}_{\text{ascent}}$ and $\mathcal{L}_{\text{descent}}$ drive the CCBF  to satisfy the gradient constraint (\ref{eq:cbf-def-cond-ascent}). \ihabold{Specifically, $\mathcal{L}_{\text{descent}}$ penalizes violations of (2) during training. While it’s hard to determine violations of (2) inside the safe
and unsafe sets, we can infer the violations whenever a trajectory crosses from a state labeled as safe to one that is labeled as unsafe. Moreover, the loss term $\mathcal{L}_{\text{lip}}$ drives  the  CCBF to be smooth.} 
Finally, $\mathcal{L}_c$ drives the CCBF to be conservative.

The term inside the sum in $\mathcal{L}_{\text{c}}$ represents the soft maximum of the CCBF values of the sampled states as well as the next state appearing in the training trajectory. The temperature parameter $\tau$ controls the smoothness of the soft maximum. As $\tau$ decreases, the soft maximum approaches the actual maximum.
When $\tau = 1$,  $\mathcal{L}_{\text{c}}$ 
is similar to a loss term appearing in the objective function of learning the $Q$-function for a learned policy $\pi_k$ in a variant of CQL described in equation (4) of~\cite{kumar2020conservative}.  That variant corresponds to choosing the sampling distribution $\mu$ in equation~(\ref{eq:th3.1_update}) in Section~\ref{sec:offline_RL_prelims} of our paper to be the one that maximizes the argument of the argmin in the presence of a regularizer $\mathcal{R}(\mu)$, i.e., 
\begin{align}
\label{eq:optimal_sampling_policy}
\mu &\gets   \text{argmax}_{\mu} \alpha \, \mathbb{E}_{s\sim \mathcal{D},\,a\sim\mu(a|s)}[Q(s,a)] \nonumber \\
&\quad + \tfrac{1}{2}\,\mathbb{E}_{s,a\sim \mathcal{D}}\Big[ Q(s,a) - \hat{B}^{\pi_k}\hat{Q}^k(s,a)\Big]^2 + \mathcal{R}(\mu),
\end{align}
where $\pi_k$ is the policy obtained by (\ref{eq:optimal_sampling_policy}) at iteration $k-1$, $\mathcal{R}(\mu) = -D_{\text{KL}}(\mu,\rho)$, $D_{\text{KL}}$ is the KL-divergence, and $\rho$ is the uniform distribution over the actions. Thus, $\mu$  would be the regularized  maximizer of the  iterate $\hat{Q}^k$. Substituting that $\mu$ in equation~(\ref{eq:th3.1_update}) in Section~\ref{sec:offline_RL_prelims} results in:
\begin{align}
\hat{Q}^{k+1} &\;\leftarrow\; \arg\min_Q \; \alpha \, \mathbb{E}_{s\sim \mathcal{D}}[\log \sum_a \text{exp}(Q(s,a))] \nonumber \\
&\quad + \tfrac{1}{2}\,\mathbb{E}_{s,a\sim \mathcal{D}}\Big[ Q(s,a) - \hat{B}^{\pi_k}\hat{Q}^k(s,a)\Big]^2,
\label{eq:th3.1_update_new}
\end{align}  
 as shown in equation (4) in \cite{kumar2020conservative}.
In our setting, $\mathcal{L}_{\text{CCBF}}$ is analogous to the objective function in (\ref{eq:th3.1_update_new}) where  the $Q$-function is replaced with the barrier function $B_\phi$, the Bellman error term is replaced with the terms of $\mathcal{L}_{\text{CCBF}}$ besides $\mathcal{L}_c$ which penalize violations of the CBF conditions similar to how the Bellman error penalizes violations of the Bellman equation,  and 
 the sampling distribution for the actions in $V_x$ in $\mathcal{L}_c$ is chosen to be the policy that maximizes safety (the barrier values) with a regularizer in  the form of the distance (KL-divergence) to the uniform distribution.  Formalizing this connection is an interesting future direction.  

\section{Case Studies}

In this section, we describe our experimental setup and present our results and key observations.

\subsection{Experimental setup}

We evaluate our approach on four environments:
(1) \textbf{2D navigation}: an agent must navigate to a goal position while avoiding a static circular obstacle,
(2) \textbf{Vision-based navigation}: similar to 2D navigation but using bird-eye-view image observations,
(3) \textbf{SafetyHopper} and (4) \textbf{SafetySwimmer}: locomotion tasks from the safety gymnasium benchmark~\cite{ji2023safety}, where agents must move as quickly as possible while adhering to velocity constraints. 
In each experiment, we present the results of rolling out the policy under different conditions: with no safety filter and with a NCBF-, an iDBF-, and a CCBF-based filter. We compare these methods with FISOR \cite{zheng2024safe} in the safety gymnasium tasks. \ihabold{When training the NCBFs, iDBFs, and CCBFs, we use the same model architecture, the same set of common hyperparameters, and the same dynamics model. The only hyperparameters which differ between the different models are those specific to the different methods. Moreover, we train the NCBFs and CCBFs on the same datasets consisting of both safe and unsafe trajectories, whereas we train the iDBFs only on the safe states from safe trajectories in the dataset (also excluding the safe states in the unsafe trajectories), as we do for training BC-Safe policies. }

\paragraph{2D navigation}  
The agent has 2D single-integrator dynamics, has actuation limits that constrain the maximum speed in each dimension to be $v_\mathit{max}$, and has to reach a goal position while avoiding a static circular obstacle. Note that the agent can stop at any state  by choosing zero control. Thus, the complement of the set of states that define the obstacle is a controlled forward invariant set. Consequently, we label a state as safe if the Euclidean distance \( d \) to the center of the obstacle is greater than or equal to \(r + \epsilon \), where \( r \) is the obstacle's radius and $\epsilon$ is a small positive constant. States with \( d \leq r \) are labeled as unsafe, and states in between are not used for training. \ihabold{The small $\epsilon$ margin improves training stability, as has been observed in other works, e.g.,~\cite{qin2022sablas}. 
}

\paragraph{Vision-based navigation}  
This setup extends the 2D navigation task to a vision-based setting, where \(64 \times 64 \times 3\) RGB observations representing a top-down view of the environment are given instead of the  2D position.

\paragraph{Safety gymnasium locomotion}

We test on SafetyHopper (11D state, 3D control) and SafetySwimmer (8D state, 2D control) from safety gymnasium \cite{ji2023safety}. Agents earn rewards for forward motion but incur costs if velocity exceeds user-defined thresholds. \ihabold{We picked these two tasks as challenging ones evident by the results in Table \ref{table:exp34} which show that FISOR and other safe offline RL algorithms fail to balance the safety and task performance in them.
We 
labeled states as unsafe if they exceeded the maximum velocity constraint, and labeled them as safe otherwise. Thus, we only labeled states in the failure set as unsafe and assumed that all other states in the dataset belong to a controlled forward invariant set. 
We also explored another strategy: in trajectories that eventually enter the failure set, we ignore the states outside the failure set and label only the states inside it (i.e., those with velocity exceeding the user-defined limit) as unsafe. States in trajectories that never enter the failure set over their entire duration are instead labeled as safe.
However, this 
approach reduced the number of trajectories to sample safe states from to 135 out of 1686 trajectories for Swimmer and 130 out of 2240 for Hopper and led to poor safety-preserving performance. We summarize the key results of that experiment (detailed results in Table \ref{table:swimmer_hopper_conservative_labeling_results}) and refer the readers to Section \ref{sec:results} for definitions of normalized cost (C) and reward (R): 
In Swimmer, averaged across all nominal controllers, using CCBF as a safety filter resulted in C=1.46 and R=0.31 (compared to C=1.49 and R=0.32 for nominal controllers without safety filter), outperforming NCBF (C: 2.38, R: 0.26).
In the case of Hopper, both NCBF and CCBF performed poorly, degrading the performance of BC and BC-Safe, while
maintaining the same performance for COptiDICE. BEAR-L and BCQ-L simulations terminated after only 5-30 steps because these algorithms, when paired with the CCBF and NCBF, produced policies that quickly drove the agent into terminal states such as falling, unstable torso angles, or extreme joint positions.
Importantly, CCBF consistently outperformed NCBF under both labeling strategies, but both NCBF and CCBF achieved significantly better safety-preserving performance when trained using the full DSRL dataset compared to the trajectory-filtered approach for labeling safe states, emphasizing the importance of more data, even if it means potentially mislabeling few states.
}

\subsection{Results}
\label{sec:results}

We discuss two main observations: (1) CCBF-based filters significantly improve safety while minimally affecting performance, 
 (2) CCBF is easier to train and more robust to hyperparameters changes than baselines. Then, we discuss the results o.ion studies that help us isolate the effect of our conservative loss term $\mathcal{L}_c$. We report the results for the 2D navigation and vision-based navigation scenarios in Table~\ref{tab:exp12} using the metrics: \textbf{success \%}, the percentage of episodes in which the agent successfully reached the goal; and \textbf{collision \%}, the percentage of episodes with at least one collision. 
We report our results for the safety gymnasium tasks in Table \ref{table:exp34} using the metrics: \textbf{normalized reward} and \textbf{normalized cost}, following the evaluation criteria in the safe offline RL literature~\cite{dsrl, fu2020d4rl}.
\ihabold{
 For both the Swimmer and the Hopper scenarios, they can can be modeled as CMDPs. 
 We assume that the transition functions 
 are 
deterministic and in the form of the control-affine dynamics in (\ref{eq:latent_dynamics}). Each episode results in a trajectory $\sigma := \{x_1, u_1, r_1, c_1, \dots, x_T, u_T, r_T, c_T\}$. 
The normalized reward and cost returns of a trajectory $\sigma$ generated using policy $\pi$  are defined as follows: 
$
R_{\text{normalized}}^\pi := \frac{R_\sigma^\pi - R_{\text{min}}}{R_{\text{max}} - R_{\text{min}}} \times 100$ and $ C_{\text{normalized}}^\pi := \frac{C_\sigma^\pi}{\kappa }$,
where $R_\sigma^\pi = \sum_t r_t$ denotes the reward return of  $\sigma$, $C_\sigma^\pi = \sum_t c_t$ denotes its cost return, and $R_{\text{min}}$ and $R_{\text{max}}$ denote the maximum and minimum reward returns of all the trajectories in the dataset as defined and computed in DSRL \cite{dsrl}. In all benchmark tasks, the 
cost $c_t$ is zero if the state $x_{t}$ is safe and one if it is unsafe. 
The Appendix contains the detailed tables of experimental results from our ablation studies, along with the hyperparameters used across all experiments.
}

\begin{table}[ht]
\centering
\setlength{\tabcolsep}{3pt}
\renewcommand{\arraystretch}{1.1}
\scriptsize
\caption{Controller evaluation results (S=Success\%, C=Collision\%). Results averaged across 500 runs (2D) and 50 runs (vision-based). {\color[HTML]{006400}\textbf{Green}}: Safest agent with the highest average success \% across all controllers.}
\begin{tabular}{l cc cc cc cc}
\toprule
\textbf{Policy} & \multicolumn{2}{c}{\textbf{None}} & \multicolumn{2}{c}{\textbf{NCBF}} & \multicolumn{2}{c}{\textbf{CCBF}} & \multicolumn{2}{c}{\textbf{iDBF}}\\
\cmidrule(lr){2-3}\cmidrule(lr){4-5}\cmidrule(lr){6-7}\cmidrule(lr){8-9}
 & S & C & S & C & S & C & S & C\\
\midrule
\multicolumn{9}{c}{\textit{2D Navigation}}\\
\midrule
PD controller & 100.0 & 90.4 & 91.6 & 0 & 91.0 & 0 & 0 & 100.0\\
BC & 96.2 & 21.8 & 95.2 & 0 & 95.4 & 0 & 9.6 & 3.4\\
BC‑Safe & 93.6 & 6.4 & 93.8 & 0 & 92.8 & 0 & 7.8 & 0\\
\midrule
\textbf{Average} & 94.9 & 14.1 & {\color[HTML]{006400}\textbf{93.5}} & {\color[HTML]{006400}\textbf{0}} & 93.1 & 0 & 5.8 & 34.5\\
\midrule
\multicolumn{9}{c}{\textit{Vision‑based Navigation}}\\
\midrule
PD controller & 100.0 & 96.0 & 34.0 & 20.0 & 58.0 & 4.0 & 94.0 & 80.0\\
BC & 98.0 & 36.0 & 68.0 & 6.0 & 86.0 & 0 & 96.0 & 14.0\\
BC‑Safe & 92.0 & 22.0 & 78.0 & 16.0 & 94.0 & 0 & 98.0 & 10.0\\
\midrule
\textbf{Average} & 95.0 & 29.0 & 60.0 & 14.0 & {\color[HTML]{006400}\textbf{79.3}} & {\color[HTML]{006400}\textbf{1.3}} & 96.0 & 34.7\\
\bottomrule
\end{tabular}
\label{tab:exp12}
\end{table}

\begin{table}[ht]
\centering
\setlength{\tabcolsep}{3pt} 
\renewcommand{\arraystretch}{1.1} 
\tiny 
\caption{Evaluation results of normalized reward (R$\uparrow$) and normalized cost (C$\downarrow$) for different pairs of nominal controllers and safety filters. Results are averaged across three barrier models trained with each method and each evaluated on 20 runs starting from random initial states. \textbf{Bold}: Safe agents (C $<$ 1). {\color[HTML]{656565} \textbf{Gray}}: Unsafe agents (C $\geq$1). {\color[HTML]{006400}\textbf{Green}}: Safest agent with the highest average success \% across all controllers.} 

\begin{tabular}{lcccccccc}

\toprule
\multirow{2}{*}{\textbf{Policy}} & \multicolumn{2}{c}{\textbf{None}} & \multicolumn{2}{c}{\textbf{NCBF}} & \multicolumn{2}{c}{\color[HTML]{006400}\textbf{CCBF}} & \multicolumn{2}{c}{\textbf{iDBF}}  \\
\cmidrule(lr){2-3} \cmidrule(lr){4-5} \cmidrule(lr){6-7} \cmidrule(lr){8-9}
& \textbf{C$\downarrow$} & \textbf{R$\uparrow$} & \textbf{C$\downarrow$} & \textbf{R$\uparrow$} & \textbf{C$\downarrow$} & \textbf{R$\uparrow$} & \textbf{C$\downarrow$} & \textbf{R$\uparrow$} \\
\midrule
\multicolumn{9}{c}{\textit{Swimmer}} \\
\midrule
BC & {\color[HTML]{656565}2.26±0.64} & \color[HTML]{656565}0.44±0.03 & \color[HTML]{656565}5.05±2.84 & \color[HTML]{656565}0.4±0.06 & \textbf{0.94±0.18}& \textbf{0.39±0.03} & 3.80±0.31 & 0.36±0.01 \\
BC-Safe & \textbf{0.12±0.05} & \textbf{0.43±0.03} & \textbf{0.12±0.06} & \textbf{0.46±0.02} & \textbf{0.03±0.02} & \textbf{0.45±0.02} & \textbf{0.19±0.06} & \textbf{0.50±0.02} \\
COptiDICE & {\color[HTML]{656565}1.65±0.33} & \color[HTML]{656565}0.30±0.04 & {\color[HTML]{656565}1.40±0.39} & \color[HTML]{656565}0.07±0.01 & \textbf{0.19±0.08} & \textbf{0.28±0.06} & \color[HTML]{656565}11.75±2.99 & \color[HTML]{656565}0.57±0.03 \\
BEAR-L & \textbf{0.61±0.11} & \textbf{0.16±0.03} & \textbf{0.46±0.05} & \textbf{0.18±0.05} & \textbf{0.42±0.17} & \textbf{0.09±0.01} & \textbf{0.43±0.07} & \textbf{0.14±0.01} \\
BCQ-L & {\color[HTML]{656565}2.82±1.15} & \color[HTML]{656565}0.25±0.08 & \color[HTML]{656565}4.69±3.20 & \color[HTML]{656565}0.20±0.10 & {\color[HTML]{656565}1.40±0.47} & \color[HTML]{656565}0.29±0.03 & \color[HTML]{656565}9.38±1.65 & \color[HTML]{656565}0.39±0.08 \\
\midrule
\textbf{Average} & {\color[HTML]{656565}1.49} & {\color[HTML]{656565}0.32} & {\color[HTML]{656565}2.34} & {\color[HTML]{656565}0.26} & \color[HTML]{006400}\textbf{0.59} & \color[HTML]{006400}\textbf{0.3} & {\color[HTML]{656565}5.1} & {\color[HTML]{656565}0.39} \\
\midrule
FISOR & \textbf{0.01±0.01} & \textbf{-0.08±0.01} & - & - & - & - & - & - \\
\midrule
\multicolumn{9}{c}{\textit{Hopper}} \\
\midrule
BC & \textbf{0.19±0.27} & \textbf{0.04±0.02} & \textbf{0.23±0.18} & \textbf{0.04±0.01}& \textbf{0.09±0.06} & \textbf{0.04±0.01} & \color[HTML]{656565}4.71±2.26 & \color[HTML]{656565}0.33±0.24 \\ 
BC-Safe & \textbf{0.03±0.02} & \textbf{0.57±0.01} & \textbf{0.14±0.10} & \textbf{0.61±0.01} & \textbf{0.05±0.03} & \textbf{0.56±0.05} & \textbf{0.21±0.13} & \textbf{0.56±0.03} \\
COptiDICE & \textbf{0.01±0.01} & \textbf{0.18±0.01} & \textbf{0.01±0.01} & \textbf{0.17±0.01} & \textbf{0.03±0.03} & \textbf{0.17±0.01} & {\color[HTML]{656565}1.81±0.73} & \color[HTML]{656565}0.24±0.03 \\
BEAR-L & \textbf{0.37±0.01} & \textbf{0.16±0.01} & \textbf{0.27±0.09} & \textbf{0.21±0.08} & \textbf{0.20±0.06} & \textbf{0.30±0.06} & \textbf{0.34±0.02} & \textbf{0.16±0.01} \\
BCQ-L & {\color[HTML]{656565}3.09±0.34} & \color[HTML]{656565}0.50±0.04 & {\color[HTML]{656565}1.34±0.48} & \color[HTML]{656565}0.31±0.02 & {\color[HTML]{656565}1.16±0.53} & \color[HTML]{656565}0.37±0.05 & \color[HTML]{656565}3.82±0.17 & \color[HTML]{656565}0.46±0.20 \\
\midrule
\textbf{Average} & \textbf{0.74} & \textbf{0.29} & \textbf{0.39} & \textbf{0.27} & \color[HTML]{006400}{\textbf{0.31}} & \color[HTML]{006400}\textbf{0.28} & {\color[HTML]{656565}2.18} & {\color[HTML]{656565}0.35} \\
\midrule
FISOR & \textbf{0.03±0.04} & \textbf{0.08±0.05} & - & - & - & - & - & - \\
\bottomrule
\end{tabular}
\label{table:exp34}
\end{table}

\paragraph{CCBF-based filters  improve safety while minimally affecting reward returns}
Observing the results in Table \ref{tab:exp12} for the 2D navigation task, we can see that NCBF and CCBF result in similar success percentages and both result in zero collisions, outperforming iDBF. Trajectories taken when using CCBF and NCBF as a safety filter are presented in Figure \ref{fig:NCBF-CCBF-iDBF-trajectories-visualization}.

In the vision-based navigation environment, CCBF outperforms both NCBF and iDBF by achieving better success and collision percentages using any nominal controller. Unlike the case of the 2D navigation scenario, iDBF generally performs better than NCBF. The improved performance of iDBF can be attributed to the naturally increased distance between training trajectories, which reduces the likelihood of mislabeling nearby images as safe or unsafe as the sampled images nearby a training trajectory will not overlap with the images visited by other training trajectories. 
In contrast, the training trajectories in the 2D environment are more closely spaced,
resulting in the sampled states by iDBF, which are labeled as unsafe, overlapping with the states of training trajectories which are labeled as safe.

In the Swimmer environment, CCBF reduces the average cost over all nominal controllers by approximately 60\% with around a 6\% drop in average reward.
In contrast, both iDBF and NCBF increase the cost.
In the Hopper environment, both CCBF and iDBF exhibit similar behavior seen in the Swimmer environment, whereas NCBF reduces the average cost by 47\% with only a 7\% drop in average reward, which differs from its performance in the Swimmer environment.
These results show that CCBF consistently reduces cost  while minimally impacting  reward. On the other hand, NCBF’s results vary between environments, performing poorly in Swimmer but well in Hopper, possibly because  Swimmer  results in a larger distribution shift during deployment.  

That said, FISOR results in low costs in the Swimmer and Hopper tasks (comparable to CCBF paired with BC-safe in the Swimmer task and with COptiDICE in the Hopper task). However, FISOR yields much lower rewards in both environments: -0.08 in Swimmer compared to 0.45 in the case of CCBF paired with BC-Safe, and 0.08 in Hopper compared to 0.17 when CCBF is paired with COptiDICE.

It is worth noting that the policies used during evaluation in Table \ref{table:exp34} are distinct from those used to collect the DSRL dataset. This demonstrates that CCBFs, despite being trained on offline data, can effectively improve safety for new policies without noticeably affecting task performance.

iDBF performs well with the BC-Safe controller, improving safety in vision-based navigation and maintaining comparable levels in Swimmer and Hopper. This is expected since iDBF’s approach labels states as unsafe when they correspond to low-probability actions under BC-Safe. However, with other controllers such as COptiDICE, performance degrades, likely because iDBF steers them toward regions that BC-Safe deems safe but that are suboptimal if the dataset trajectories are not expert (reward-maximizing).

To further investigate iDBF's performance, we conducted extensive ablation experiments by varying the model size and the density threshold $p$ for selecting OOD actions. We tested medium-sized models (4 layers, 128 neurons) versus large models (5 layers, 400 neurons) across different density thresholds ($p \in \{1, 0.1, 10^{-4}, 10^{-8}\}$). For medium-sized models, average costs increased substantially across all thresholds compared to not using any safety filter: 
average Swimmer costs rose from 1.49 to 4.38-6.92, while Hopper costs increased from 0.74 to 1.02-2.50. Large models generally showed improvement with average costs ranging from 3.85-6.63 for Swimmer and 0.77-1.90 for Hopper, demonstrating better performance than medium-sized models across most density thresholds. However, across all configurations and model sizes, iDBF's performance consistently deteriorated and underperformed compared to nominal controllers, and to CCBF and NCBF methods that leverage actual unsafe trajectories in the offline dataset. Detailed results of these experiments can be found in Appendix \ref{app:idbf_size} and \ref{app:extreme_labeling}. 

\paragraph{CCBFs is easier to train and more robust to changes in hyperparameters}
In all experiments, CCBFs consistently distinguished safe and unsafe states when assigning $w_c$ to values between 0 and 0.5. However, tuning $w_c$ (typically between 0.5 and 1) was needed for  CCBF to sufficiently disincentivize unseen actions, as shown in Figure \ref{fig:NCBF-CCBF-iDBF-trajectories-visualization}.
On the other hand, iDBF’s performance was more sensitive to the density threshold $p$ that is used to classify sampled actions as OOD. We also had to tune the hyperparameters $w_{\text{safe}}$, $w_{\text{unsafe}}$, $\varepsilon_{\text{safe}}$, and  $\varepsilon_{\text{unsafe}}$ to  obtain an iDBF that does not assign a single value to all states.


\paragraph{Isolating the effect of our conservativeness-inducing  loss term $\mathcal{L}_c$}

\begin{figure}[ht]
\centering
\includegraphics[width=\columnwidth]{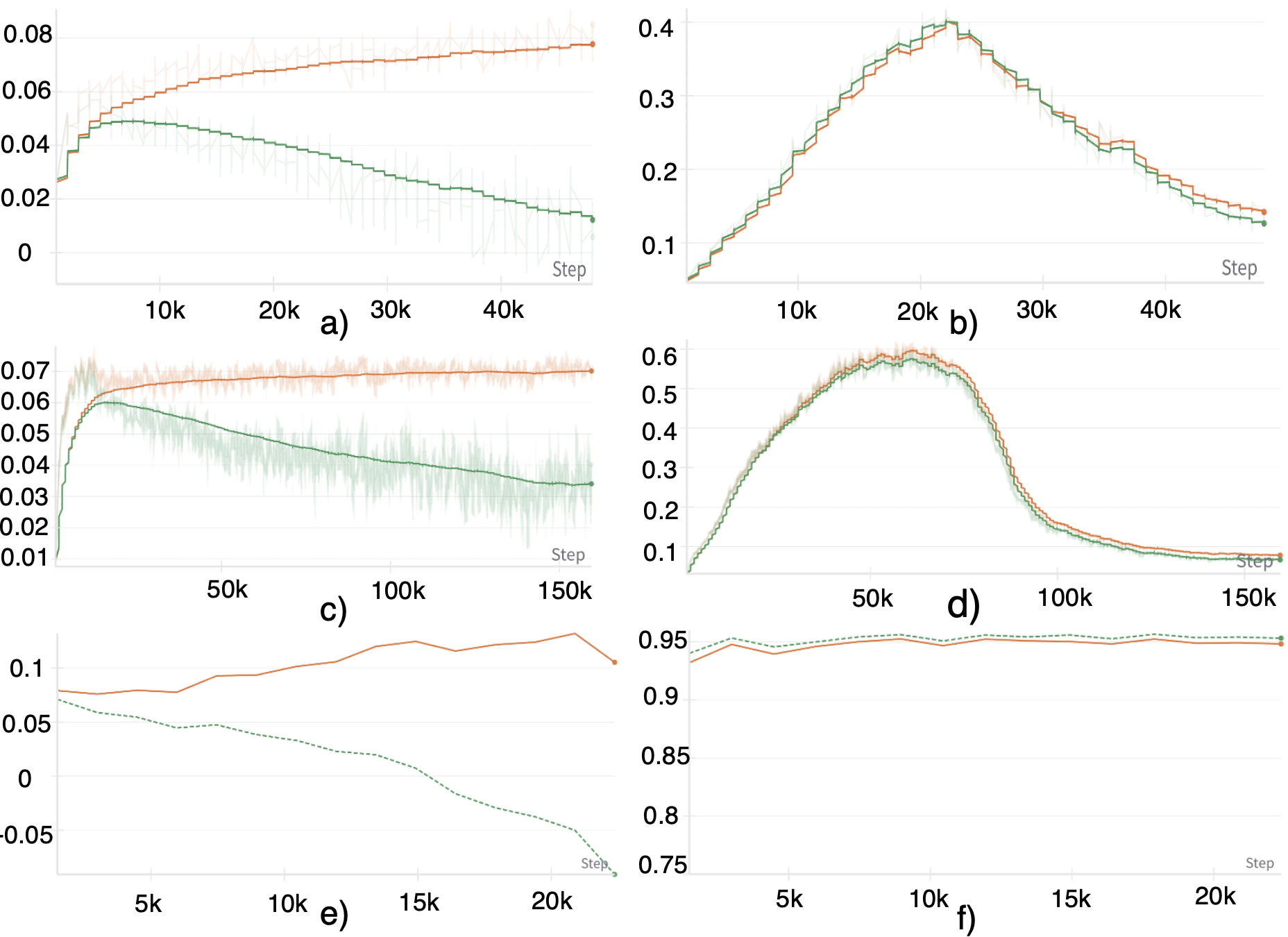}
\caption{Variation of the mean of the barrier values of states reached by taking either dataset actions (\textcolor{orange}{orange}) or randomly sampled actions (\textcolor{ForestGreen}{green}) starting from safe states in unseen test trajectories, plotted against the number of training steps. Results are shown for CCBF trained with $w_c=0.5$: (a) Swimmer, (c) Hopper, and (e) vision-based navigation, where dataset actions are consistently evaluated as safer than random actions. For NCBF: (b) Swimmer, (d) Hopper, and (f) vision-based navigation, where this distinction is not observed and the two curves largely overlap.}

\label{fig:logs_for_b(x')}
\end{figure}

We conducted an experiment where we trained CCBFs with $w_c \in \{0, 0.5, 1.0\}$, for the vision-based navigation, Swimmer, and Hopper environments. Note that when $w_c=0$, our approach is equivalent to training a NCBF. We split the dataset trajectories into training and testing ones and only train on the former. 
For each safe state in the test trajectories, we evaluated the CCBF at the next state along the trajectory as well as at a set of states obtained by taking random actions from that same state, following the procedure used to compute $\mathcal{L}_c$. 
First, we found that increasing $w_c$ decreases the CCBF values of the states reached by taking both the actions in the test set and the actions randomly sampled, as expected. Second, increasing $w_c$ enlarges the gap between the values of the CCBFs at the states in the test set compared to the ones reached by the random actions. This clearly shows that the learned CCBFs generalize well to distinguish actions taken by the policies generating the data and random ones and consider the states resulting from the latter as less safe. In contrast, we found that NCBFs consistently assign similar values for the randomly-reached states and the ones in the test set, also as expected. We present examples of our results in Figure \ref{fig:logs_for_b(x')} and show more detailed ones in Figures \ref{fig:hopper-barrier-comparison}, \ref{fig:swimmer-barrier-comparison}, and  \ref{fig:vision-barrier-comparison}.

Next, we studied whether other conservativeness-inducing loss terms 
would result in similar improvement in performance compared to the NCBF approach. 
We denote the first candidate loss term, which replaces $\mathcal{L}_c$, by $\mathcal{L}_{\text{unsafe}}'$ and its associate weight by $w_{\text{unsafe}}'$. 
The term $\mathcal{L}_{\text{unsafe}}'$ is equivalent $\mathcal{L}_{\text{unsafe}}$ evaluated at the states sampled in the same manner when computing $\mathcal{L}_c$, i.e., treating the states reached  by randomly-sampled actions similarly to unsafe ones. 
We call this new approach CCBF$^*$.  It exhibits similar properties to CCBF when comparing the values it assigns for the OOD and InD states in new trajectories and performs better than iDBF and NCBF, but under performs compared to CCBF. We summarise the main results below and refer readers to Appendix \ref{app:ccbf*_full_table} for details. For Swimmer, the average results across all controllers are:
CCBF$^*$ with $w_c^* = 0.5$ ($C=1.33$, $R=0.34$) 
and $w_c^* = 1$ ($C=2.44$, $R=0.32$). 
For Hopper, the averages are:
CCBF$^*$ with $w_c^* = 0.5$ ($C=0.37$, $R=0.31$), 
and $w_c^* = 1$ ($C=0.91$, $R=0.26$). 
We note that the learned CCBF* resulting from $w_c^*=1$ under performs the one resulting from $w_c^*=0.5$. This potentially occurs because $w_c^*=1$ penalizes OOD states as much as actual unsafe states. Since randomly reached states are likely (but not necessarily) OOD, this aligns with findings in \cite{li2022data} stating that treating OOD and unsafe actions similarly degrades performance. 
 As a remark, in both CCBF and CCBF*, increasing $w_c$ and $w_c^*$ above the value of one can result in over-conservative safety filters which lead to low rewards. 

 From these two experiments, we conclude that CCBF's and CCBF*'s  conservativeness-inducing loss terms enable learning safety filters that not only filter actions for safety, but also for staying in-distribution. Our experiments show that CCBF outperforms CCBF*, but further evaluation is needed to claim that one of these approaches is consistently better than the other, which we leave for future work. 
 
\section{Conclusion}
We introduced Conservative Control Barrier Functions (CCBFs), neural safety filters trained using offline data. The core idea is to add a loss term that prevents over-estimation of the safety of state-action pairs not seen in the dataset.
When reference controllers and safety filters are trained offline, CCBFs, even when paired with reference controllers that were not used to generate the offline dataset, outperform baselines and result in closed-loop behaviors that better preserve safety while minimally impacting task performance. Our approach in its present form does not offer formal guarantees for safety maintenance or OOD state avoidance, and deriving ones would be exciting future work. Also, evaluating our method on maintaining the safety of real robots would also be needed for further validation. 
\bibliography{references}
\newpage

\clearpage

\appendix
\section*{}


\subsection{Experimental details for 2D navigation environment}\label{app:2dtask}
\subsubsection{Dataset generation and implementation details}\label{app:2d_task_imp_details}
We generated 2,000 diverse trajectories using a CBF-QP policy that combines a nominal PD controller with a manually-designed CBF. The nominal controller is defined as \( \mathbf{u}_{\text{nom}} = K_p (\mathbf{x}_g - \mathbf{x}) + K_d \frac{\mathbf{u}_{\text{nom}}}{\Delta t} \), where \( \mathbf{x}_g \) is the goal position, \( \Delta t = 0.1 \) seconds is the timestep (10 Hz frequency), and \( K_p, K_d \) are the proportional and derivative gains, respectively. The constraints in the QP are  \( \dot{h}(\mathbf{x}) + \alpha h(\mathbf{x}) \geq 0 \), where \( h(\mathbf{x}) = \| \mathbf{x} - \mathbf{x}_{\text{obs}} \|^2 - r^2 \) is the CBF, and  $-v_{\text{max}} \leq  v_x\leq v_{\text{max}}$ and  $-v_{\text{max}} \leq  v_y\leq v_{\text{max}}$, where $v_{\text{max}} = 3$. Each trajectory is initialized from a random state. We gradually increase obstacle radius $r$ as the dataset generation progresses. Initially, we set r to $0.01$ and increment it by 0.2 up to 5, updating every 25000 time steps, where the time step count is cumulative over trajectories. The episode ends after 200 timesteps or when the agent reaches the goal, defined as \(\|\mathbf{x} - \mathbf{x}_{\text{g}}\|^2 < \delta\), where \(\delta\) is 0.5. This setup generates a diverse dataset of safe and unsafe trajectories. We train the CCBF as described in Section \ref{sec:learn_cbf}, but we do not include the terms $\mathcal{L}_{\text{descent}}$  and $\mathcal{L}_{\text{lip}}$ in this scenario.

During evaluation, each episode starts from a randomly sampled initial state within the square \([-18, 5]^2 \). 
We solve the QP without actuation constraints, and then scale the control input as \( \mathbf{u}_{\mathrm{scaled}} = \frac{\mathbf{u}}{\|\mathbf{u}\|_2}\, v_{\max} \) to ensure it satisfies the actuation constraints. 


    
\begin{table}[H]
\centering
\caption{Hyperparameters for NCBF, CCBF, iDBF in the 2D navigation task}
\label{tab:cbf_second_env_hyperparameters}
\begin{tabular}{lc}
\toprule
\textbf{Parameter} & \textbf{Value} \\
\midrule
Number of hidden layers & 3\\
Hidden dimension & 128\\
Batch size & 128 \\
Learning rate & $1 \times 10^{-4}$ \\
$\varepsilon_{\text{safe}}$& 0.2 \\
$\varepsilon_{\text{unsafe}}$ & 0.2 \\
$w_{\text{safe}}$ & 1.0 \\
$w_{\text{unsafe}}$ & 1.2 \\
$w_{\text{ascent}}$ & 1.0 \\

$p$ & $1 \times 10^{-8}$ \\
$w_c$ & 0.5\\
$\tau$ & 0.7\\
Optimizer & Adam \\
Time step ($\texttt{dt}$) & 0.1 \\
\bottomrule
\end{tabular}
\end{table}

\subsection{Experimental details for vision-based navigation environment}\label{app:2dtask}
\subsubsection{Dataset generation and implementation details}\label{app:vision_based_task}
Using the same setup as in the 2D navigation task, we set $\delta=2$ and collect 3,000 trajectories with  \(64 \times 64 \times 3\) RGB image observations.  They  are a mix of safe and unsafe trajectories, generated by varying the obstacle radius and initial positions as before, with the maximum episode length being 20 seconds (200 timesteps, 10Hz controller). The collected image-based trajectories are then used to learn latent dynamics, as described in Section~\ref{sec:learn_latent_dynamics}. An observation is considered safe if the Euclidean distance \( d \) to the obstacle is greater than or equal to \(r + \epsilon \), where $\epsilon=3$, otherwise it is unsafe. We train the CCBF as described in Section \ref{sec:learn_cbf}, but do not include $\mathcal{L}_{\text{lip}}$.


\begin{table}[H]
\centering
\caption{Hyperparameters for NCBF, CCBF, and iDBF in the vision-based navigation task}
\label{tab:hyperparameters}
\begin{tabular}{lc}
\toprule
\textbf{Parameter} & \textbf{Value} \\
\midrule
Latent dimension size& 4 \\
Action dimension size  & 2 \\
Hidden dimension  & 128 \\
Number of hidden layers  & 3 \\
Batch size  & 256 \\
Learning rate  & $1 \times 10^{-4}$ \\
$\varepsilon_{\text{safe}}$ & 0.08 \\
$\varepsilon_{\text{unsafe}}$ & 0.15 \\
$\varepsilon_{\text{ascent}}$ & 0.02 \\
$\varepsilon_{\text{descent}}$ & 0.02 \\
$w_{\text{ascent}}$& 2.0 \\
$w_{\text{decent}}$& 1.0 \\
$w_{\text{safe}}$& 1.0 \\
$w_{\text{unsafe}}$& 1.1 \\
$p$ & $1 \times 10^{-6}$ \\
$w_c$ & 1\\
$\tau$ & 0.7\\
Optimizer & Adam \\
Time step ($\texttt{dt}$) & 0.1 \\
\bottomrule
\end{tabular}
\end{table}

\begin{table}[ht]
\centering
\setlength{\tabcolsep}{3pt} 
\renewcommand{\arraystretch}{1.1} 
\tiny 
\caption{Hyperparameters for latent dynamics in the vision-based navigation task}
\label{tab:vae_dynamics_architecture}
\begin{tabular}{lc}
\toprule
\textbf{Parameters} & \textbf{Value} \\
\midrule
Encoder CNN layers &
\begin{tabular}[t]{@{}l@{}}
Conv2d(3$\rightarrow$32) \\
Conv2d(32$\rightarrow$64) \\
Conv2d(64$\rightarrow$128) \\
Flatten \\
Linear(8192$\rightarrow$400) \\
Linear(400$\rightarrow$4) (latent state)
\end{tabular}
\\[0.5em]
Decoder CNN layers &
\begin{tabular}[t]{@{}l@{}}
Linear(4$\rightarrow$8192) \\
Unflatten (128, 8, 8) \\
ConvTranspose2d(128$\rightarrow$64) \\
ConvTranspose2d(64$\rightarrow$32) \\
ConvTranspose2d(32$\rightarrow$3) \\
Sigmoid activation (output $\in [0,1]$)
\end{tabular}
\\

Input channels  & 3 \\
Latent dimension size & 4 \\
Hidden dimension (Dynamics ) & 400 \\
Number of hidden layers (Dynamics) & 3 \\
Learning rate (Dynamics)  & $1 \times 10^{-4}$ \\
Learning rate (encoder, decoder) & $1 \times 10^{-4}$ \\

Batch size  & 32 \\
$\lambda_1$ & 1.0 \\
$\lambda_2$ & 1.0 \\
$\lambda_3$ & 0.5 \\
Optimizer & Adam \\
Time step ($\texttt{dt}$) & 0.1 \\
\bottomrule
\end{tabular}
\end{table}

 \label{sec:swimmer_and_hopper_env}
 \subsection{Implementation and experimental details in the Swimmer and Hopper tasks}\label{app:swimmer_hopper_implementation_details}
 We first train a dynamics model as described in Section \ref{sec:learn_latent_dynamics}, but instead of employing an autoencoder as in the vision-based task to map observations to latent states, we learn the dynamics directly in safety gymnasium’s native state space. Then, we train the CCBF as described in Section \ref{sec:learn_cbf}, but do not include $\mathcal{L}_{\text{descent}}$. We train the nominal controllers using the offline RL algorithms (COptiDICE \cite{lee2022coptidice}, BEAR-L \cite{xu2022constraints,kumar2019stabilizing}, BCQ-L \cite{xu2022constraints,fujimoto2019off}) and using BC variants as implemented in DSRL~\cite{dsrl}. 
 During evaluation, if the QP was infeasible, we select the closest point in \(\mathcal{U}\) to the constraint boundary.

 \begin{table}[h]
\centering
\caption{Hyperparameters for the NCBF and dynamics model in the safety gymnasium environments}
\label{tab:cbf_safetygym_hyperparameters}
\begin{tabular}{lc}
\toprule
\textbf{Parameter} & \textbf{Value} \\
\midrule
hidden dimension (CBF) & 128 \\
Number of hidden layers (CBF) & 4 \\
Learning rate (CBF) & $1 \times 10^{-5}$ \\
$\varepsilon_{\text{safe}}$ & 0.08 \\
$\varepsilon_{\text{unsafe}}$ & 0.15 \\
$\varepsilon_{\text{ascent}}$& 0.02\\
$w_{\text{ascent}}$ & 2\\ 
$w_{\text{safe}}$ & 1.0 \\
$w_{\text{unsafe}}$ & 1.2 \\
$w_{\text{lip}}$ & 1.0 \\

Batch size& 256 \\
Hidden dimension (dynamics) & 128 \\
Number of hidden layers (dynamics) & 4 \\
Learning rate (dynamics) & $1 \times 10^{-4}$ \\
Optimizer & Adam \\
Time step ($\texttt{dt}$) & 0.1 \\
\bottomrule
\end{tabular}
\end{table}
For each of Swimmer and Hopper, we selected the best hyperparameters across three different models when reporting averaged results. 
In the Hopper environment, we choose the best $w_c$ and $\tau$ parameters for the CCBF models as 
$(0.1, 1)$, $(0.05, 1)$, and $(0.5, 0.7)$, respectively. 
For the iDBF models in Hopper, we select $p$ values of $1 \times 10^{-10}$, $1 \times 10^{-5}$, and $1 \times 10^{-8}$, respectively. Similarly, in the Swimmer environment, we select the best $w_c$ and $\tau$ parameters for the three CCBF models as 
$(1,0.5)$, $(1, 1)$, and $(1, 0.7)$, respectively. 
For the iDBF models in Swimmer, we select $p$ values of $1 \times 10^{-6}$, $1 \times 10^{-9}$, and $1 \times 10^{-12}$, respectively. We also train the safety filters in Swimmer and Hopper for 15k and 50k steps respectively and choose the best checkpoints.

\subsection{Closed loop evaluation results for CCBF and NCBF in the trajectory-filtered approach for labeling safe state}
\label{app:labels_swimmer_hopper}

\begin{table}[ht]
\centering
\setlength{\tabcolsep}{3pt} 
\renewcommand{\arraystretch}{1.1} 
\tiny 
\caption{Evaluation of normalized reward (R$\uparrow$) and normalized cost (C$\downarrow$) across different policies using NCBF and CCBF under the trajectory-filtered approach for labeling safe state. The $\uparrow$ symbol denotes that the higher reward, the better. The $\downarrow$ symbol denotes that the lower cost, the better.}
\centering
\begin{tabular}{lcc cc cc}
\toprule
\multirow{2}{*}{\textbf{Policy}} & \multicolumn{2}{c}{\textbf{None}} & \multicolumn{2}{c}{\textbf{NCBF}} & \multicolumn{2}{c}{\textbf{CCBF}} \\
\cmidrule(lr){2-3} \cmidrule(lr){4-5} \cmidrule(lr){6-7}
& \textbf{C$\downarrow$} & \textbf{R$\uparrow$} & \textbf{c$\downarrow$} & \textbf{R$\uparrow$} & \textbf{C$\downarrow$} & \textbf{R$\uparrow$} \\
\midrule
\multicolumn{7}{c}{\textit{Swimmer}} \\
\midrule
BC & 2.26±0.64 & 0.44±0.03 & 3.61±1.10 & 0.31±0.05 & 0.88±0.15 & 0.33±0.02 \\
BC‑Safe & 0.12±0.05 & 0.43±0.03 & 0.16±0.02 & 0.40±0.01 & 0.12±0.02 & 0.43±0.04 \\
COptiDICE & 1.65±0.33 & 0.30±0.04 & 6.08±0.95 & 0.37±0.08 & 3.20±0.57 & 0.31±0.02 \\
BEAR‑L & 0.61±0.11 & 0.16±0.03 & 0.47±0.08 & 0.18±0.02 & 0.47±0.17 & 0.14±0.01 \\
BCQ‑L & 2.82±1.15 & 0.25±0.08 & 1.56±0.30 & 0.05±0.01 & 2.62±1.20 & 0.36±0.09 \\
\midrule
\textbf{Average} & 1.49 & 0.32 & 2.38 & 0.26 & 1.46 & 0.31 \\
\midrule
\multicolumn{7}{c}{\textit{Hopper}} \\
\midrule
BC & 0.19±0.27 & 0.04±0.02 & 0.39±0.02 & 0.02±0.00 & 0.41±0.08 & 0.02±0.00 \\
BC‑Safe & 0.03±0.02 & 0.57±0.01 & 0.41±0.10 & 0.19±0.02 & 0.30±0.01 & 0.19±0.02 \\
COptiDICE & 0.01±0.01 & 0.18±0.01 & 0.03±0.01 & 0.15±0.00 & 0.02±0.05 & 0.13±0.01 \\
BEAR‑L & - & - & - & - & - & - \\
BCQ‑L & - & - & - & - & - & - \\
\midrule
\textbf{Average} & 0.08 & 0.26 & 0.28 & 0.12 & 0.24 & 0.11 \\
\bottomrule
\end{tabular}
\label{table:swimmer_hopper_conservative_labeling_results}
\end{table}

\section{Ablation studies}
\label{app:ablation_studies}

\subsection{Influence of model size and density threshold \(p\) on performance of iDBF}
\label{app:idbf_size}

To remain faithful to the iDBF methodology and address whether model size may be causing the sub-par closed-loop performance observed in the safety gymnasium environments, we evaluated two iDBF architectures. The first is a medium-sized neural network with 4 hidden layers of 128 neurons each, and the second is a larger model with 5 hidden layers of 400 neurons each. Tables \ref{tab:ablation1} and \ref{tab:ablation2} report the corresponding results.

For the Hopper environment, our results show that increasing the model size reduces the normalized cost across all tested density thresholds \(p\), but it also reduces the reward. This indicates that a larger model improves iDBF’s ability to classify safe states from the training data and differentiate them from unsafe states generated via the BC-safe policy. In contrast, the Swimmer environment exhibits mixed effects: a larger model reduces both cost and reward when \(p=10^{-8}\), but leads to an increased cost and improved reward for \(p=0.1\) and \(p=10^{-4}\). This shows that while iDBF still degrades the performance relative to the nominal policy without safety filters across both model architectures, the degradation is less severe when using the larger model.

\begin{table}[ht]
\centering
\setlength{\tabcolsep}{3pt} 
\renewcommand{\arraystretch}{1.1} 
\tiny 
\caption{Evaluation results of normalized reward (R$\uparrow$) and normalized cost (C$\downarrow$) across different policies and a \textbf{medium-sized} iDBF model trained with 4 hidden layers, 128 neurons each, and with varying density threshold $p$. The $\uparrow$ symbol denotes that the higher reward, the better. The $\downarrow$ symbol denotes that the lower cost, the better. }
\centering
\begin{tabular}{lcc cc cc cc}
\toprule
\multirow{2}{*}{\textbf{Policy}} & \multicolumn{2}{c}{\textbf{None}} & \multicolumn{2}{c}{\textbf{iDBF ($p=0.1$)}} & \multicolumn{2}{c}{\textbf{iDBF ($p=10^{-4}$)}} & \multicolumn{2}{c}{\textbf{iDBF ($p=10^{-8}$)}} \\
\cmidrule(lr){2-3} \cmidrule(lr){4-5} \cmidrule(lr){6-7} \cmidrule(lr){8-9}
 & \textbf{C$\downarrow$} & \textbf{R$\uparrow$} & \textbf{C$\downarrow$} & \textbf{R$\uparrow$} & \textbf{C$\downarrow$} & \textbf{R$\uparrow$} & \textbf{C$\downarrow$} & \textbf{R$\uparrow$} \\
\midrule
\multicolumn{9}{c}{\textit{Swimmer}} \\
\midrule
BC        & 2.26±0.64    & 0.44±0.03    & 3.24±1.41   & 0.31±0.05   & 5.82±1.13   & 0.41±0.04   & 5.45±1.54   & 0.45±0.03 \\
BC‑Safe   & 0.12±0.05    & 0.43±0.03    & 0.19±0.03   & 0.42±0.01   & 0.22±0.03   & 0.47±0.02   & 0.27±0.03   & 0.50±0.01 \\
COptiDICE & 1.65±0.33    & 0.30±0.04    & 8.06±0.49   & 0.44±0.02   & 10.43±0.55  & 0.53±0.03   & 11.71±0.87  & 0.58±0.01 \\
BEAR‑L    & 0.61±0.11    & 0.16±0.03    & 0.57±0.23   & 0.15±0.03   & 0.62±0.10   & 0.13±0.02   & 0.40±0.08   & 0.14±0.02 \\
BCQ‑L     & 2.82±1.15    & 0.25±0.08    & 9.83±2.06   & 0.49±0.07   & 9.53±1.52   & 0.39±0.02   & 11.07±0.90  & 0.46±0.02 \\
\midrule
\textbf{Average}  & 1.49         & 0.32         & 4.38        & 0.36        & 5.32        & 0.39        & 5.78        & 0.43 \\
\midrule
\multicolumn{9}{c}{\textit{Hopper}} \\
\midrule
BC        & 0.19±0.27    & 0.04±0.02    & 3.87±0.18   & 0.33±0.06   & 7.49±1.26   & 0.63±0.13   & 2.28±0.88   & 0.25±0.06 \\
BC‑Safe   & 0.03±0.02    & 0.57±0.01    & 0.17±0.07   & 0.54±0.04   & 0.35±0.09   & 0.58±0.02   & 0.25±0.06   & 0.60±0.01 \\
COptiDICE & 0.01±0.01    & 0.18±0.01    & 1.15±0.23   & 0.23±0.01   & 0.17±0.10   & 0.19±0.01   & 0.40±0.07   & 0.18±0.01 \\
BEAR‑L    & 0.37±0.01    & 0.16±0.01    & 0.35±0.01   & 0.16±0.00   & 0.35±0.01   & 0.16±0.00   & 0.35±0.01   & 0.16±0.00 \\
BCQ‑L     & 3.09±0.34    & 0.50±0.04    & 4.46±0.18   & 0.55±0.05   & 4.16±0.29   & 0.55±0.01   & 3.20±0.47   & 0.55±0.01 \\
\midrule
\textbf{Average}  & 0.74         & 0.29         & 2.00        & 0.36        & 2.50        & 0.42        & 1.30        & 0.35 \\
\bottomrule
\end{tabular}
\label{tab:ablation1}
\end{table}

\begin{table}[ht]
\centering
\setlength{\tabcolsep}{3pt} 
\renewcommand{\arraystretch}{1.1} 
\tiny 
\caption{Evaluation results of normalized reward (R$\uparrow$) and normalized cost (C$\downarrow$) across different policies and a \textbf{large-sized} iDBF model trained with 5 hidden layers, 400 neurons each, and with varying density threshold $p$. The $\uparrow$ symbol denotes that the higher reward, the better. The $\downarrow$ symbol denotes that the lower cost, the better. }
\centering
\begin{tabular}{lcc cc cc cc}
\toprule
\multirow{2}{*}{\textbf{Policy}} & \multicolumn{2}{c}{\textbf{None}} & \multicolumn{2}{c}{\textbf{iDBF ($p=0.1$)}} & \multicolumn{2}{c}{\textbf{iDBF ($p=10^{-4}$)}} & \multicolumn{2}{c}{\textbf{iDBF ($p=10^{-8}$)}} \\
\cmidrule(lr){2-3} \cmidrule(lr){4-5} \cmidrule(lr){6-7} \cmidrule(lr){8-9}
 & \textbf{C$\downarrow$} & \textbf{R$\uparrow$} & \textbf{C$\downarrow$} & \textbf{R$\uparrow$} & \textbf{C$\downarrow$} & \textbf{R$\uparrow$} & \textbf{C$\downarrow$} & \textbf{R$\uparrow$} \\
\midrule
\multicolumn{9}{c}{\textit{Swimmer}} \\
\midrule
BC        & 2.26±0.64  & 0.44±0.03  & 6.79±1.48  & 0.42±0.03  & 8.41±0.70  & 0.44±0.05  & 4.00±1.22  & 0.54±0.04 \\
BC‑Safe   & 0.12±0.05  & 0.43±0.03  & 0.19±0.01  & 0.43±0.02  & 0.32±0.03  & 0.50±0.00  & 0.47±0.09  & 0.49±0.00 \\
COptiDICE & 1.65±0.33  & 0.30±0.04  & 10.09±0.82 & 0.52±0.02  & 14.13±0.23 & 0.44±0.01  & 5.36±0.28  & 0.54±0.00 \\
BEAR‑L    & 0.61±0.11  & 0.16±0.03  & 0.55±0.15  & 0.11±0.03  & 0.48±0.05  & 0.14±0.01  & 0.53±0.05  & 0.13±0.02 \\
BCQ‑L     & 2.82±1.15  & 0.25±0.08  & 9.46±1.28  & 0.41±0.07  & 12.47±1.79 & 0.49±0.06  & 8.90±0.72  & 0.56±0.03 \\
\midrule
\textbf{Average}  & 1.49       & 0.32       & 5.42       & 0.38       & 7.16       & 0.40       & 3.85       & 0.45 \\
\midrule
\multicolumn{9}{c}{\textit{Hopper}} \\
\midrule
BC        & 0.19±0.27  & 0.04±0.02  & 0.91±0.48  & 0.08±0.03  & 2.56±0.53  & 0.15±0.05  & 3.34±0.53  & 0.21±0.04 \\
BC‑Safe   & 0.03±0.02  & 0.57±0.01  & 0.33±0.07  & 0.59±0.02  & 0.16±0.08  & 0.51±0.02  & 0.08±0.05  & 0.45±0.02 \\
COptiDICE & 0.01±0.01  & 0.18±0.01  & 0.21±0.04  & 0.19±0.01  & 0.06±0.06  & 0.19±0.00  & 0.01±0.01  & 0.18±0.00 \\
BEAR‑L    & 0.37±0.01  & 0.16±0.01  & 0.08±0.05  & 0.20±0.01  & 0.11±0.00  & 0.16±0.00  & 0.12±0.01  & 0.16±0.00 \\
BCQ‑L     & 3.09±0.34  & 0.50±0.04  & 2.30±0.91  & 0.13±0.05  & 2.67±0.24  & 0.22±0.02  & 2.55±0.31  & 0.19±0.05 \\
\midrule
\textbf{Average}  & 0.74       & 0.29       & 0.77       & 0.24       & 1.11       & 0.25       & 1.22       & 0.24 \\
\bottomrule
\end{tabular}
\label{tab:ablation2}
\end{table}

\subsection{Learning iDBF with density threshold $p=1$}
\label{app:extreme_labeling}

\ihabold{From each state in the dataset comprised of safe trajectories, we sampled random actions, took a one‑step forward simulation, and marked all the resulting states as unsafe. This procedure mimics the effect of choosing a very high density threshold \(p=1\) within the iDBF framework. We call this approach \textbf{Extreme iDBF}. Results are shown in Table \ref{table:ablation3}.

We tested this method with both the medium- and large-sized iDBF models to explore whether additional model capacity aids in learning better safety boundaries, given the inherent difficulty involved in this method where nearby states receive conflicting labels (safe and unsafe).

The empirical findings indicate that, for both model sizes, on average across all nominal controllers, this extreme labeling method results in poorer performance compared to the nominal policy without any safety filter. Notably, across both Swimmer and Hopper environments, pairing BEAR-L with this extreme version of iDBF achieves lower cost (with similar reward levels) than using BEAR-L without the safety filter. However, combining COptiDICE or BCQ-L with iDBF severely degrades performance, as evidenced by a significant increase in cost.}

\begin{table}[ht]
\centering
\setlength{\tabcolsep}{3pt} 
\renewcommand{\arraystretch}{1.1} 
\tiny 
\caption{Evaluation of normalized reward (R$\uparrow$) and normalized cost (C$\downarrow$) across different policies using iDBF with both \textbf{medium-} and \textbf{large-}sized models, each employing an \textbf{extreme labeling strategy (Extreme iDBF): every state reached by a one-step forward simulation with a randomly sampled action is labeled as unsafe irrespective of the density threshold $p$.} The $\uparrow$ symbol denotes that the higher reward, the better. The $\downarrow$ symbol denotes that the lower cost, the better. }
\centering

\begin{tabular}{lcc cc cc}

\toprule
\multirow{2}{*}{\textbf{Policy}} & \multicolumn{2}{c}{\textbf{None}} & \multicolumn{2}{c}{\textbf{Extreme iDBF (medium-sized)}} & \multicolumn{2}{c}{\textbf{Extreme iDBF (large-sized)}} \\
\cmidrule(lr){2-3} \cmidrule(lr){4-5} \cmidrule(lr){6-7}
 & \textbf{C$\downarrow$} & \textbf{R$\uparrow$} & \textbf{C$\downarrow$} & \textbf{R$\uparrow$} & \textbf{C$\downarrow$} & \textbf{R$\uparrow$} \\
\midrule
\multicolumn{7}{c}{\textit{Swimmer}} \\
\midrule

BC         & 2.26±0.64  & 0.44±0.03  & 9.18±1.05  & 0.55±0.01  & 5.55±1.02  & 0.46±0.05 \\
BC‑Safe    & 0.12±0.05  & 0.43±0.03  & 0.54±0.12  & 0.49±0.01  & 0.19±0.03  & 0.47±0.02 \\
COptiDICE  & 1.65±0.33  & 0.30±0.04  & 13.63±0.65 & 0.59±0.01  & 15.68±0.31 & 0.48±0.01 \\
BEAR‑L     & 0.61±0.11  & 0.16±0.03  & 0.41±0.08  & 0.15±0.02  & 0.54±0.20  & 0.15±0.02 \\
BCQ‑L      & 2.82±1.15  & 0.25±0.08  & 10.84±0.72 & 0.45±0.02  & 11.17±0.45 & 0.44±0.02 \\

\midrule
\textbf{Average} 
           & 1.49       & 0.32       & 6.92       & 0.45       & 6.63       & 0.40 \\
\midrule
\multicolumn{7}{c}{\textit{Hopper}} \\
\midrule
BC         & 0.19±0.27  & 0.04±0.02  & 2.01±0.28  & 0.21±0.07  & 4.75±0.47  & 0.17±0.02 \\
BC‑Safe    & 0.03±0.02  & 0.57±0.01  & 0.43±0.11  & 0.35±0.03  & 0.08±0.06  & 0.34±0.04 \\
COptiDICE  & 0.01±0.01  & 0.18±0.01  & 0.28±0.04  & 0.10±0.00  & 1.27±0.26  & 0.23±0.01 \\
BEAR‑L     & 0.37±0.01  & 0.16±0.01  & 0.12±0.01  & 0.15±0.00  & 0.19±0.01  & 0.16±0.00 \\
BCQ‑L      & 3.09±0.34  & 0.50±0.04  & 2.25±0.40  & 0.10±0.03  & 3.23±0.44  & 0.52±0.04 \\

\midrule
\textbf{Average} 
           & 0.74       & 0.29       & 1.02       & 0.18       & 1.90       & 0.28 \\
\bottomrule
\end{tabular}
\label{table:ablation3}
\end{table}

\ihabold{\subsection{Closed loop evaluation of CCBF* paired with different controllers}
\label{app:ccbf*_full_table}
we studied whether other conservativeness-inducing loss terms 
would result in similar improvement in performance compared to the NCBF approach. 
We denote the first candidate loss term, which replaces $\mathcal{L}_c$, by $\mathcal{L}_{\text{unsafe}}'$ and its associate weight by $w_{\text{unsafe}}'$. 
The term $\mathcal{L}_{\text{unsafe}}'$ is equivalent $\mathcal{L}_{\text{unsafe}}$ evaluated at the states sampled in the same manner when computing $\mathcal{L}_c$, i.e., treating the states reached  by randomly-sampled actions similarly to unsafe ones. 
We call this new approach CCBF$^*$.}

\begin{table}[ht]
\centering
\setlength{\tabcolsep}{3pt} 
\renewcommand{\arraystretch}{1.1} 
\tiny 
\caption{Evaluation of normalized reward (R$\uparrow$) and normalized cost (C$\downarrow$) across different policies using CCBF* with different $w_c^*$. The $\uparrow$ symbol denotes that the higher reward, the better. The $\downarrow$ symbol denotes that the lower cost, the better.}
\centering
\begin{tabular}{lcc cc cc}
\toprule
\multirow{2}{*}{\textbf{Policy}} & \multicolumn{2}{c}{\textbf{None}} & \multicolumn{2}{c}{\textbf{CCBF* ($w_c^* = 0.5$)}} & \multicolumn{2}{c}{\textbf{CCBF* ($w_c^* = 1$)}} \\
\cmidrule(lr){2-3} \cmidrule(lr){4-5} \cmidrule(lr){6-7}
& \textbf{C$\downarrow$} & \textbf{R} & \textbf{C$\downarrow$} & \textbf{R$\uparrow$} & \textbf{C$\downarrow$} & \textbf{R$\uparrow$} \\
\midrule
\multicolumn{7}{c}{\textit{Swimmer}} \\

\midrule
BC & 2.26±0.64 & 0.44±0.03 & 2.06±0.16 & 0.37±0.03 & 2.38±0.66 & 0.40±0.04 \\
BC‑Safe & 0.12±0.05 & 0.43±0.03 & 0.13±0.08 & 0.35±0.01 & 0.06±0.02 & 0.43±0.02 \\
COptiDICE & 1.65±0.33 & 0.30±0.04 & 0.96±0.42 & 0.48±0.01 & 6.01±0.82 & 0.35±0.03 \\
BEAR‑L & 0.61±0.11 & 0.16±0.03 & 0.46±0.09 & 0.15±0.02 & 0.56±0.16 & 0.17±0.02 \\
BCQ‑L & 2.82±1.15 & 0.25±0.08 & 3.03±0.50 & 0.33±0.02 & 3.20±0.91 & 0.26±0.04 \\
\midrule
\textbf{Average} & 1.49 & 0.32 & 1.33 & 0.34 & 2.44 & 0.32 \\
\midrule
\multicolumn{7}{c}{\textit{Hopper}} \\
\midrule
BC & 0.19±0.27 & 0.04±0.02 & 0.39±0.09 & 0.07±0.01 & 1.71±0.33 & 0.18±0.02 \\
BC‑Safe & 0.03±0.02 & 0.57±0.01 & 0.11±0.06 & 0.56±0.01 & 0.27±0.16 & 0.52±0.02 \\
COptiDICE & 0.01±0.01 & 0.18±0.01 & 0.08±0.03 & 0.18±0.01 & 0.16±0.05 & 0.18±0.01 \\
BEAR‑L & 0.37±0.01 & 0.16±0.01 & 0.37±0.17 & 0.39±0.01 & 0.24±0.13 & 0.26±0.00 \\
BCQ‑L & 3.09±0.34 & 0.50±0.04 & 0.92±0.28 & 0.34±0.01 & 2.15±05 & 0.18±0.04 \\
\midrule
\textbf{Average} & 0.74 & 0.29 & 0.37 & 0.31 & 0.91 & 0.26 \\
\bottomrule
\end{tabular}
\label{table:updated_results}
\end{table}

\subsection{Additional discussion of results: safety barrier performance is dependent on the nominal controller and the environment}
\label{app:further_results}
The closed-loop performance of a learned safety filter—whether based on NCBF, iDBF, or our proposed CCBF—is affected by the quality of the nominal  controller. A learned barrier function that is not formally verified will often not satisfy the CBF condition in (\ref{eq:cbf-def-cond-ascent}) for certain state-action pairs, and the nominal controller determines which actions at each state are examined by the QP solver.  This seems to have a minimal effect in low-dimensional tasks such as the 2D navigation experiment, as all nominal controllers equipped with NCBF and CCBF achieve zero collisions and high success rate using any of the nominal  controllers.

However, in the vision-based navigation experiment, the choice of nominal controller had a pronounced impact on performance. We can see in Table~\ref{tab:exp12} how  all the learned barrier functions perform poorly when paired with the goal reaching PD nominal controller .
However, when paired with BC and BC-Safe nominal controllers, both iDBF and CCBF perform well, with CCBF achieving zero collisions and high success rate.

Interestingly, controllers in the safety gymnasium environment based on BC variants and offline reinforcement learning show varied behavior when augmented with a barrier. For instance, in the Hopper environment, when paired with BEAR-L, CCBF reduces the average cost from 0.37 to 0.20 and raises the average reward from 0.16 to 0.30. In the Swimmer environment with the BC-Safe controller paired with CCBF, the cost decreases from 0.12 to 0.03 while the reward increases from 0.43 to 0.45. For other controllers paired with CCBF, the decrease in cost returns is traded with a small decrease in reward returns.

  \subsection{Comparison of the learned neural CBFs using CCBF, CCBF*, and NCBF in the Hopper, Swimmer, and vision-based navigation task}
  We evaluate and compare the neural control barrier functions learned using NCBF, CCBF, and CCBF*  across the Hopper, Swimmer, and vision-based navigation tasks. For each method, we examine how the barrier function values differ between safe and unsafe states in test trajectories, as well as states reached by taking either dataset actions or randomly sampled actions starting from safe states in the test trajectories.
\twocolumn[{%
  \centering
  \section{Comparison of the learned neural CBFs using different methods in the Hopper task}
  \label{app:graphs_hopper}
  \vspace{1em}
  \resizebox{\textwidth}{!}{%
    \begin{tabular}{cc}
      \includegraphics[width=0.48\textwidth]{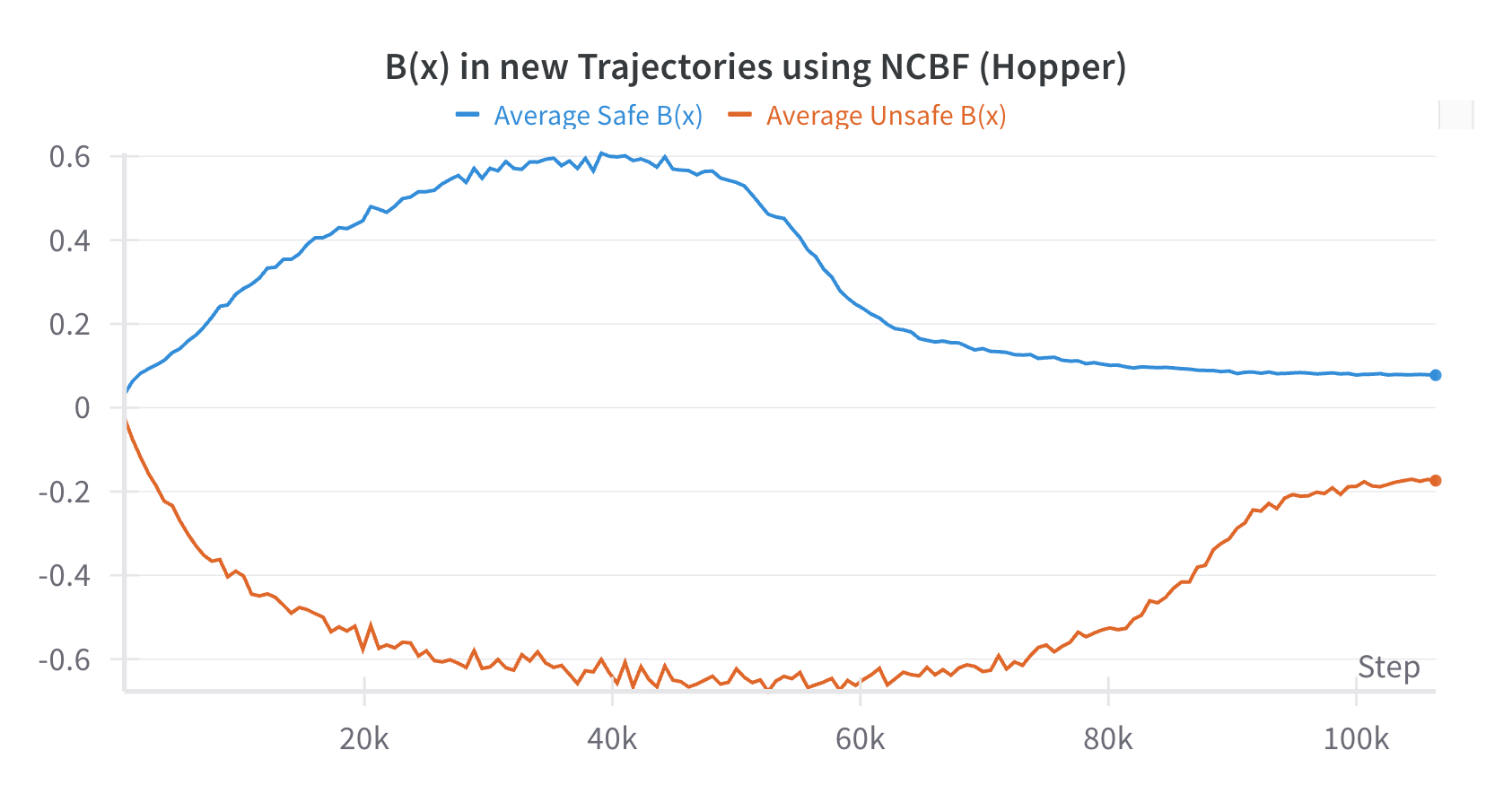} &
      \includegraphics[width=0.48\textwidth]{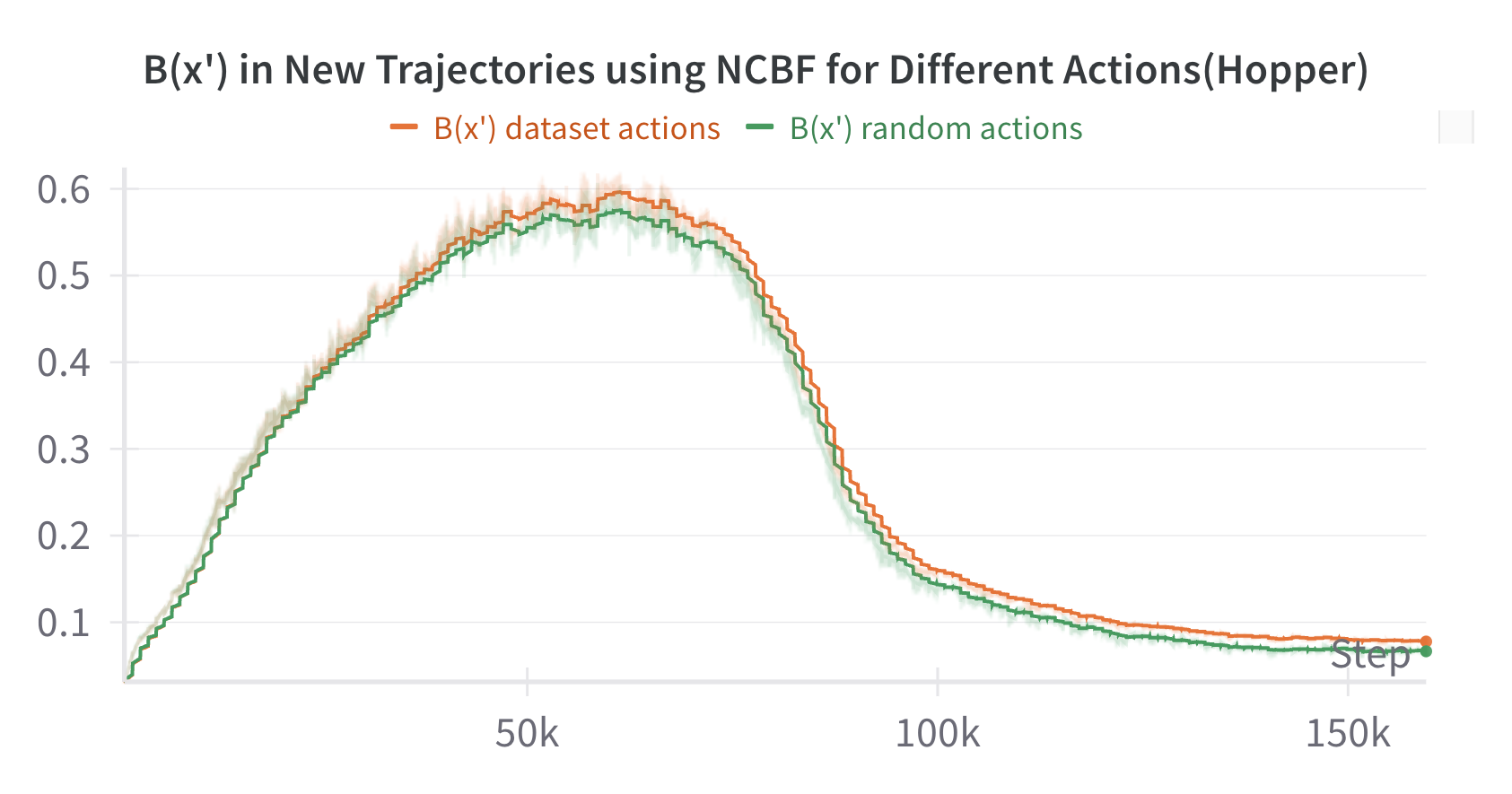} \\

      \includegraphics[width=0.48\textwidth]{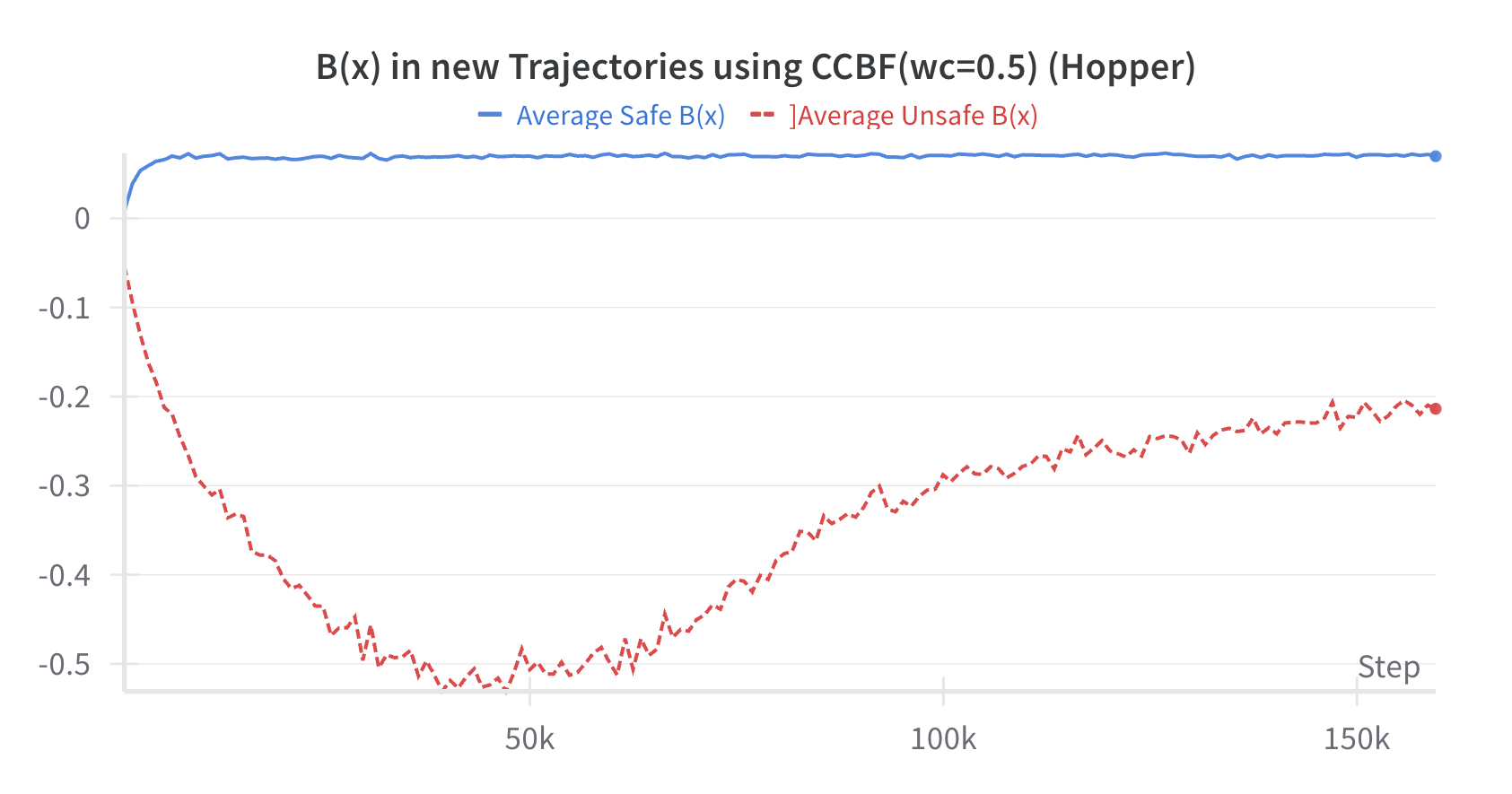} &
      \includegraphics[width=0.48\textwidth]{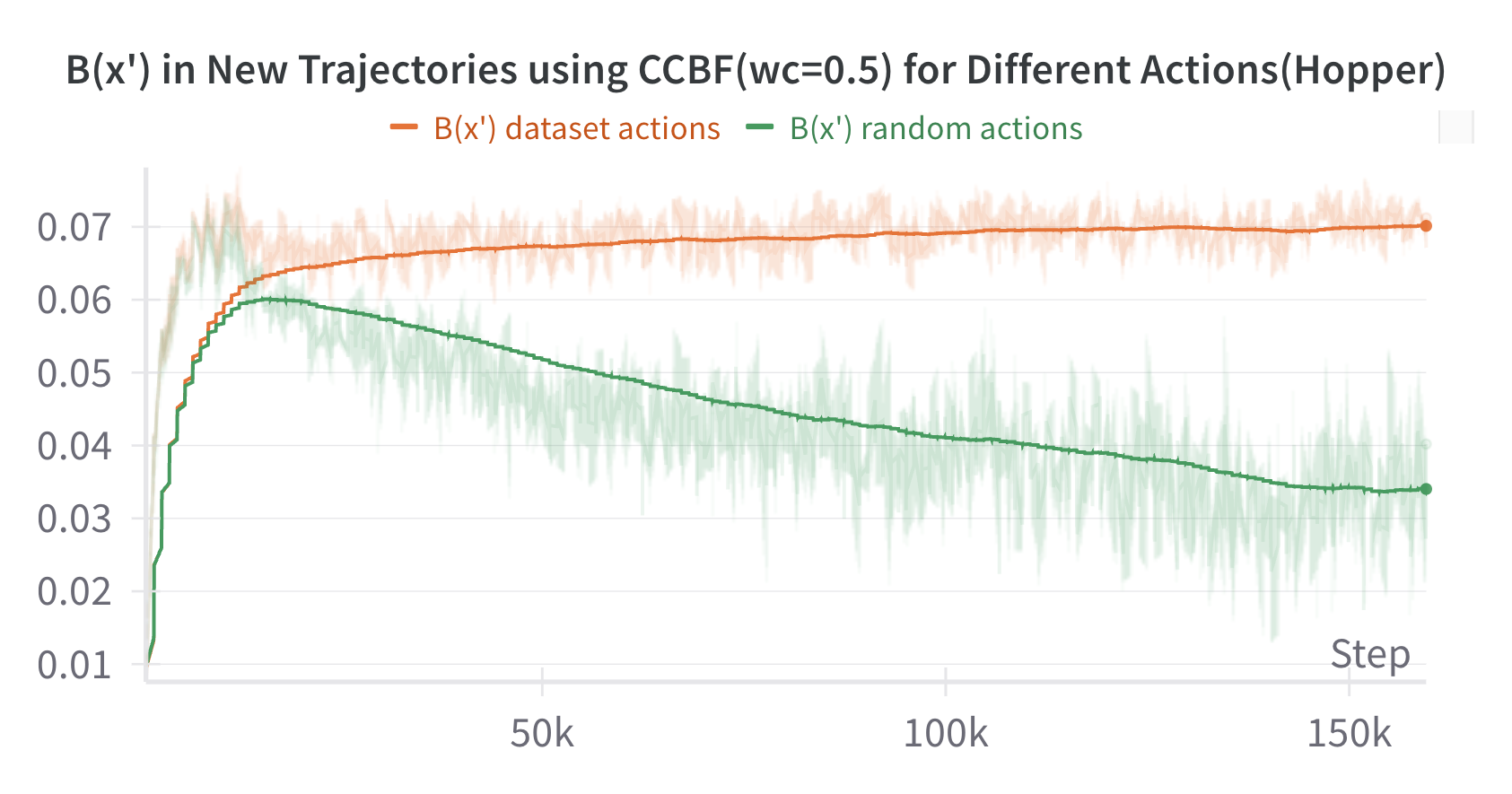} \\

      \includegraphics[width=0.48\textwidth]{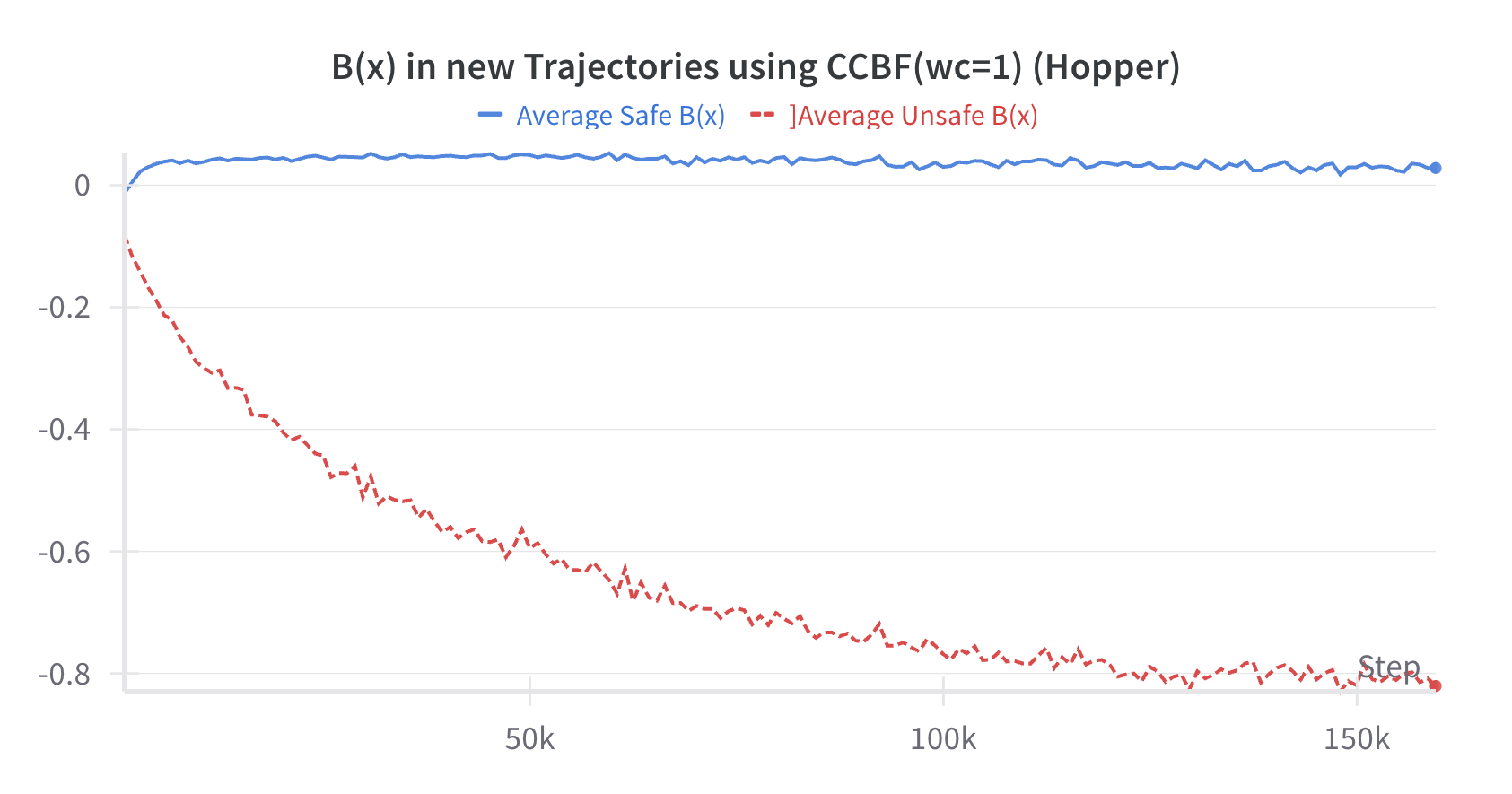} &
      \includegraphics[width=0.48\textwidth]{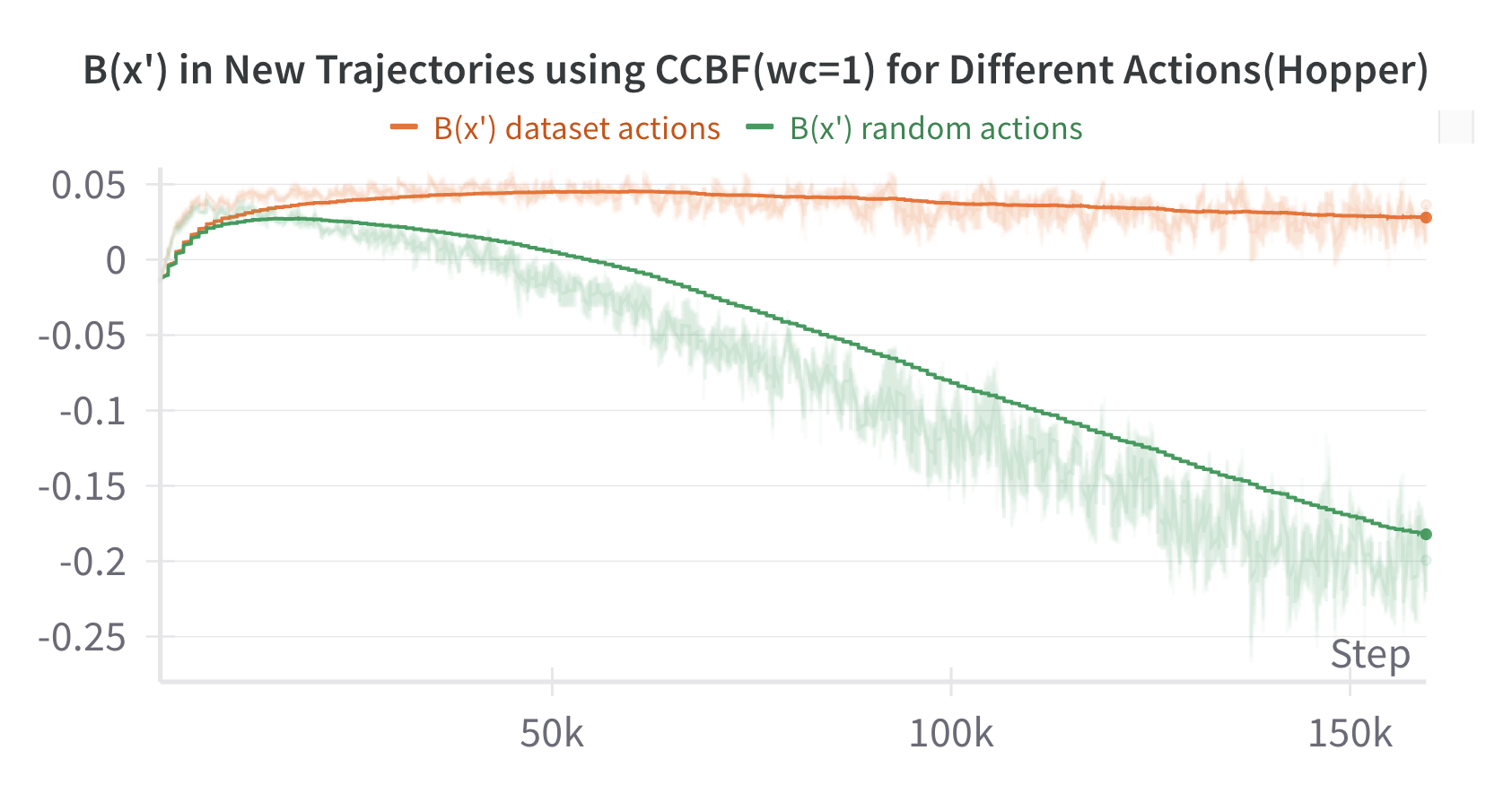} \\

      \includegraphics[width=0.48\textwidth]{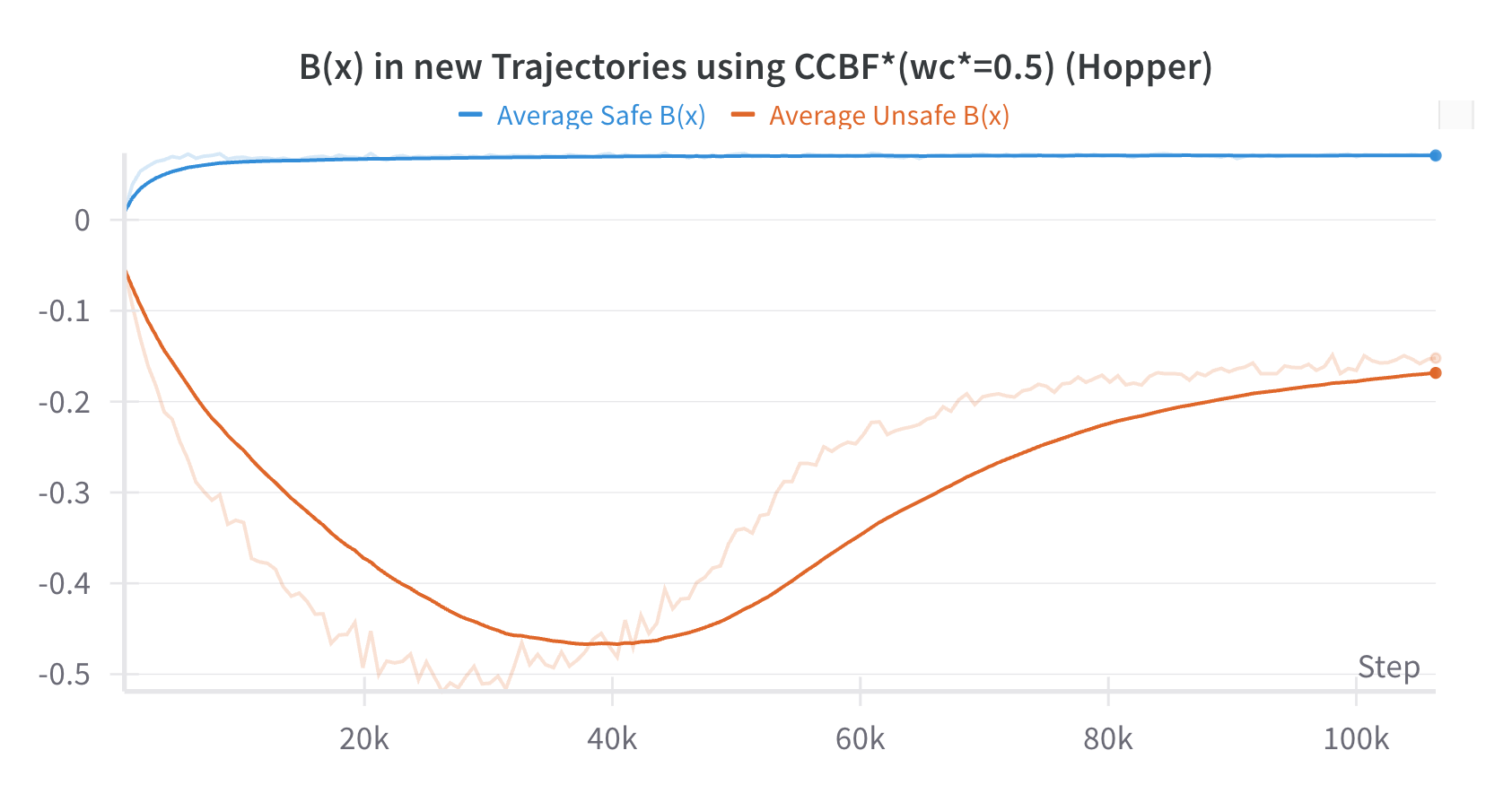} &
      \includegraphics[width=0.48\textwidth]{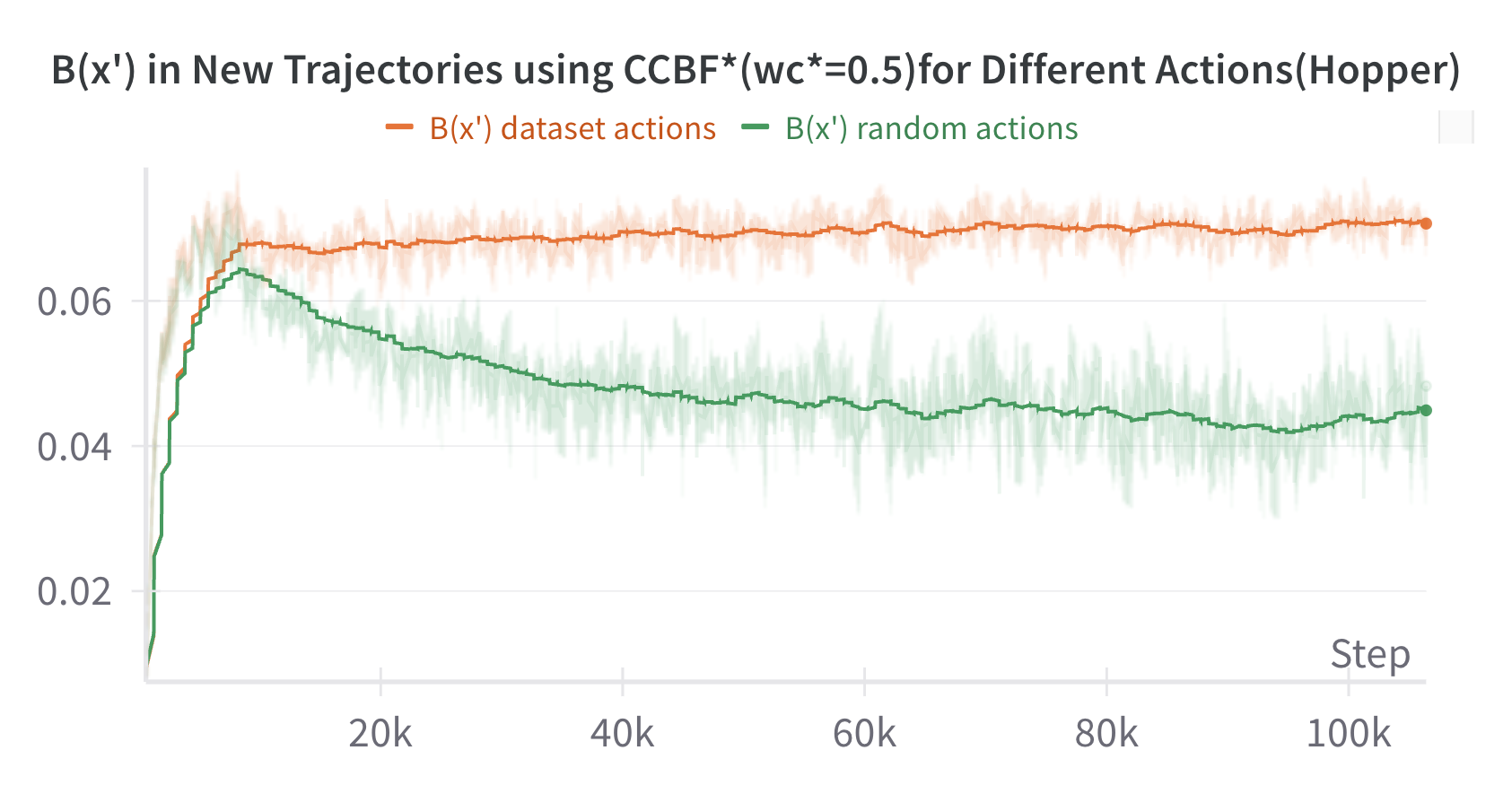} \\

      \includegraphics[width=0.48\textwidth]{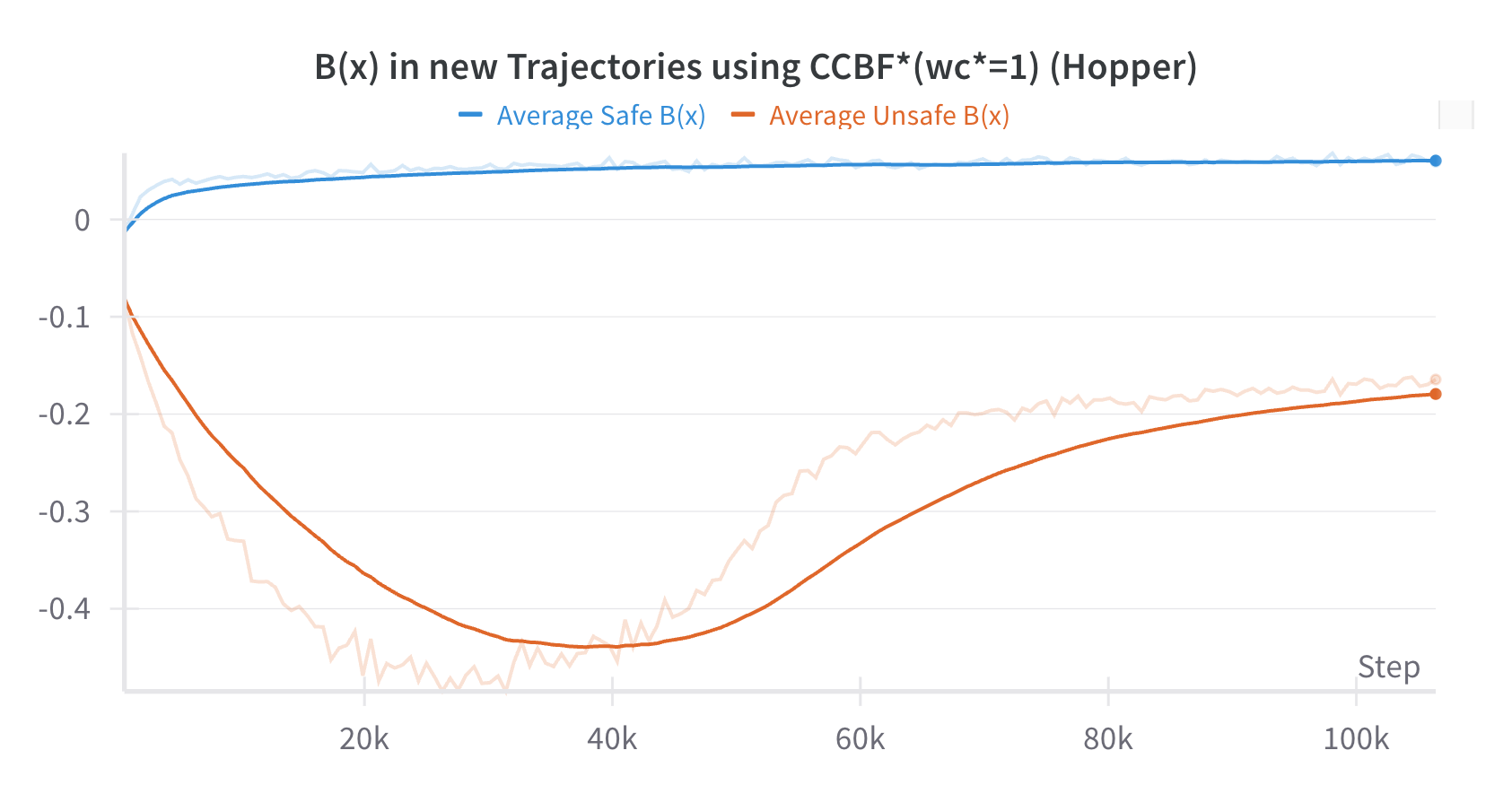} &
      \includegraphics[width=0.48\textwidth]{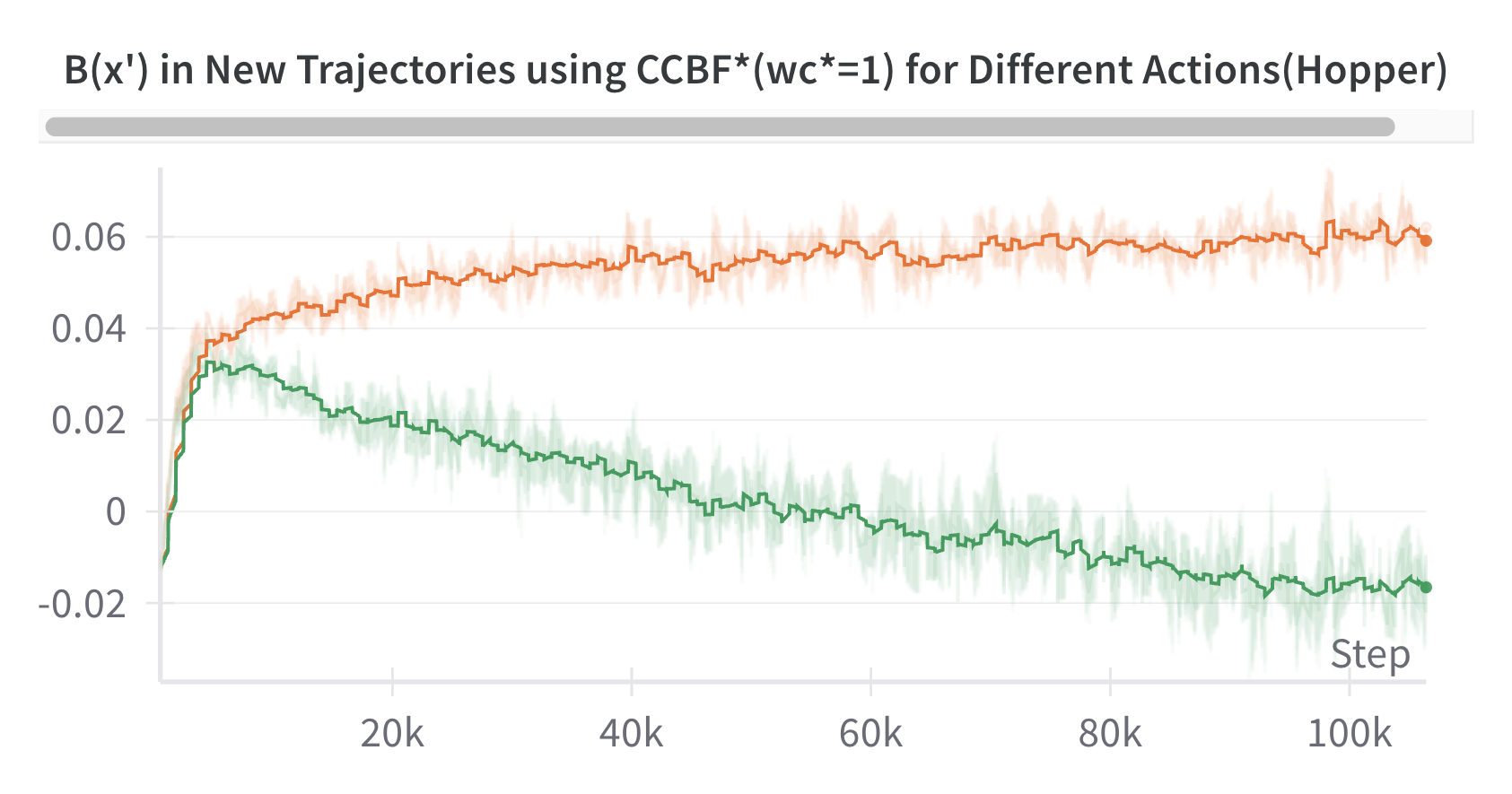} \\
    \end{tabular}
  }
  \captionof{figure}{Variation of the mean neural control barrier function values for NCBF, CCBF, and CCBF* on Hopper test trajectories. 
  \textbf{Left column:} safe vs. unsafe states in test trajectories. 
  \textbf{Right column:} states reached by taking either dataset actions or randomly sampled actions from safe states in unseen test trajectories.}
  \label{fig:hopper-barrier-comparison}
}]

\clearpage

\twocolumn[{%
  \centering
  \section{Comparison of the learned neural CBFs using different methods in the Swimmer task}
    \label{app:graphs_swimmer}
  \vspace{1em}
  \resizebox{\textwidth}{!}{%
    \begin{tabular}{cc}
      \includegraphics[width=0.48\textwidth]{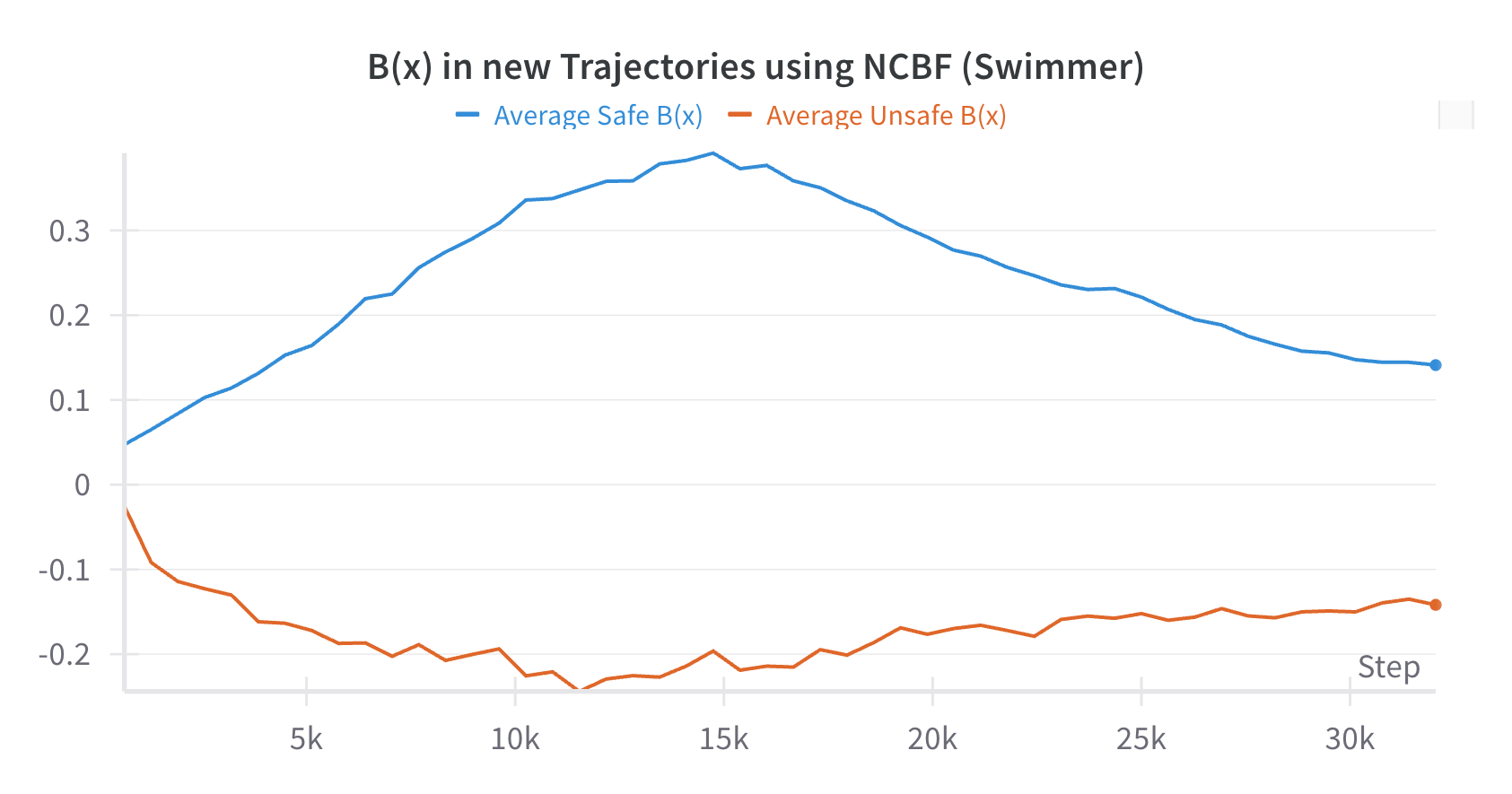} &
      \includegraphics[width=0.48\textwidth]{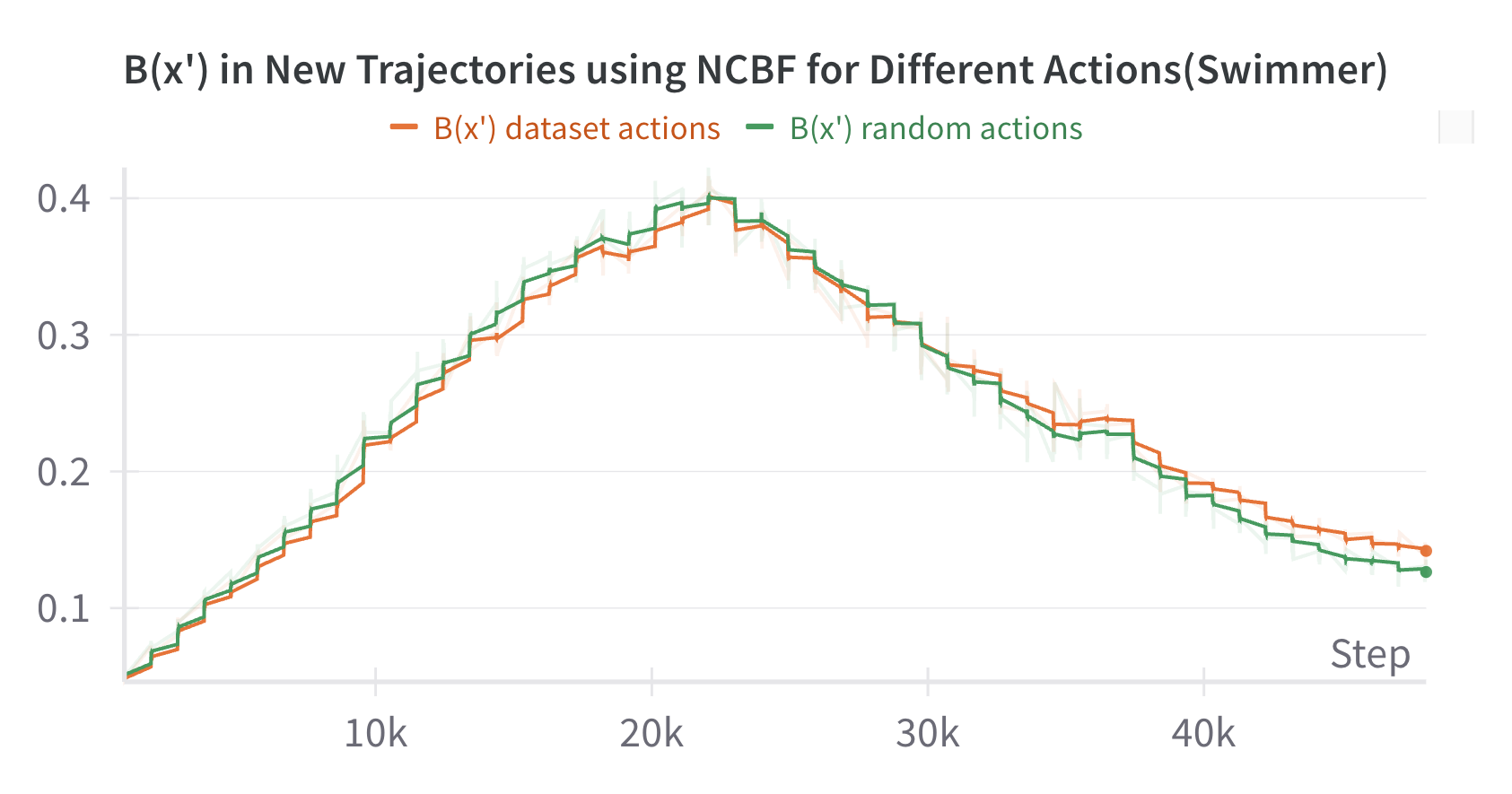} \\

      \includegraphics[width=0.48\textwidth]{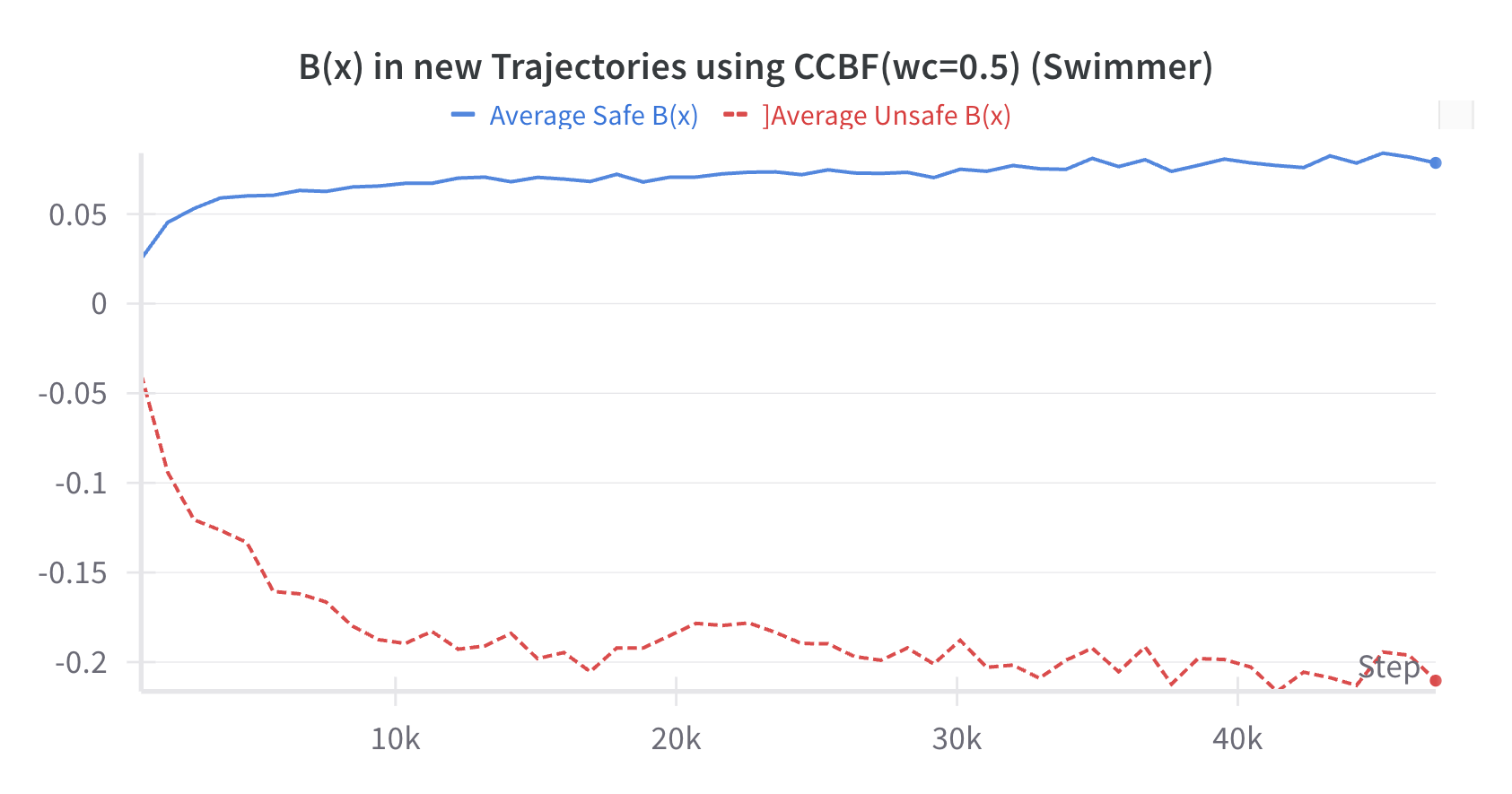} &
      \includegraphics[width=0.48\textwidth]{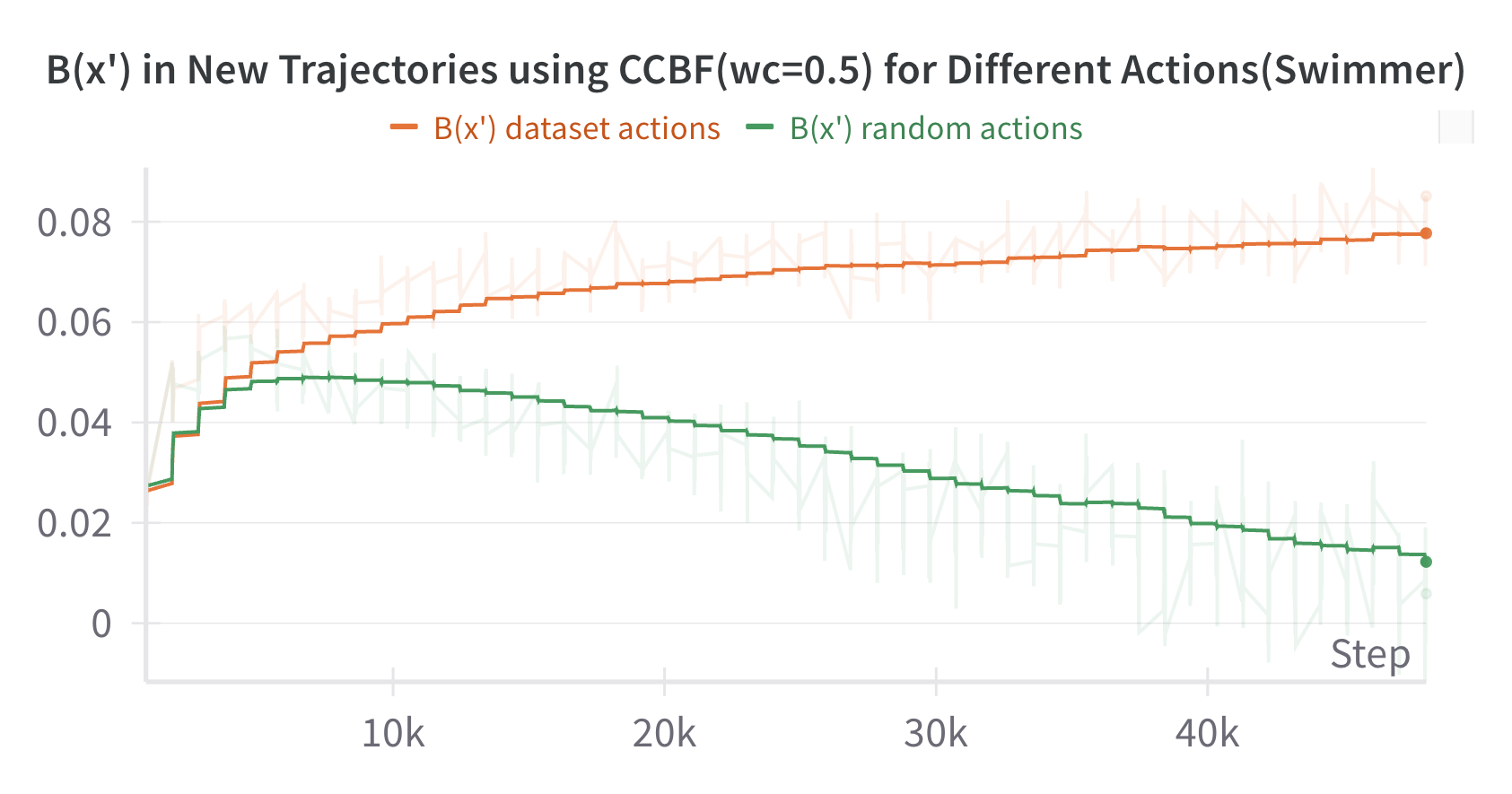} \\

      \includegraphics[width=0.48\textwidth]{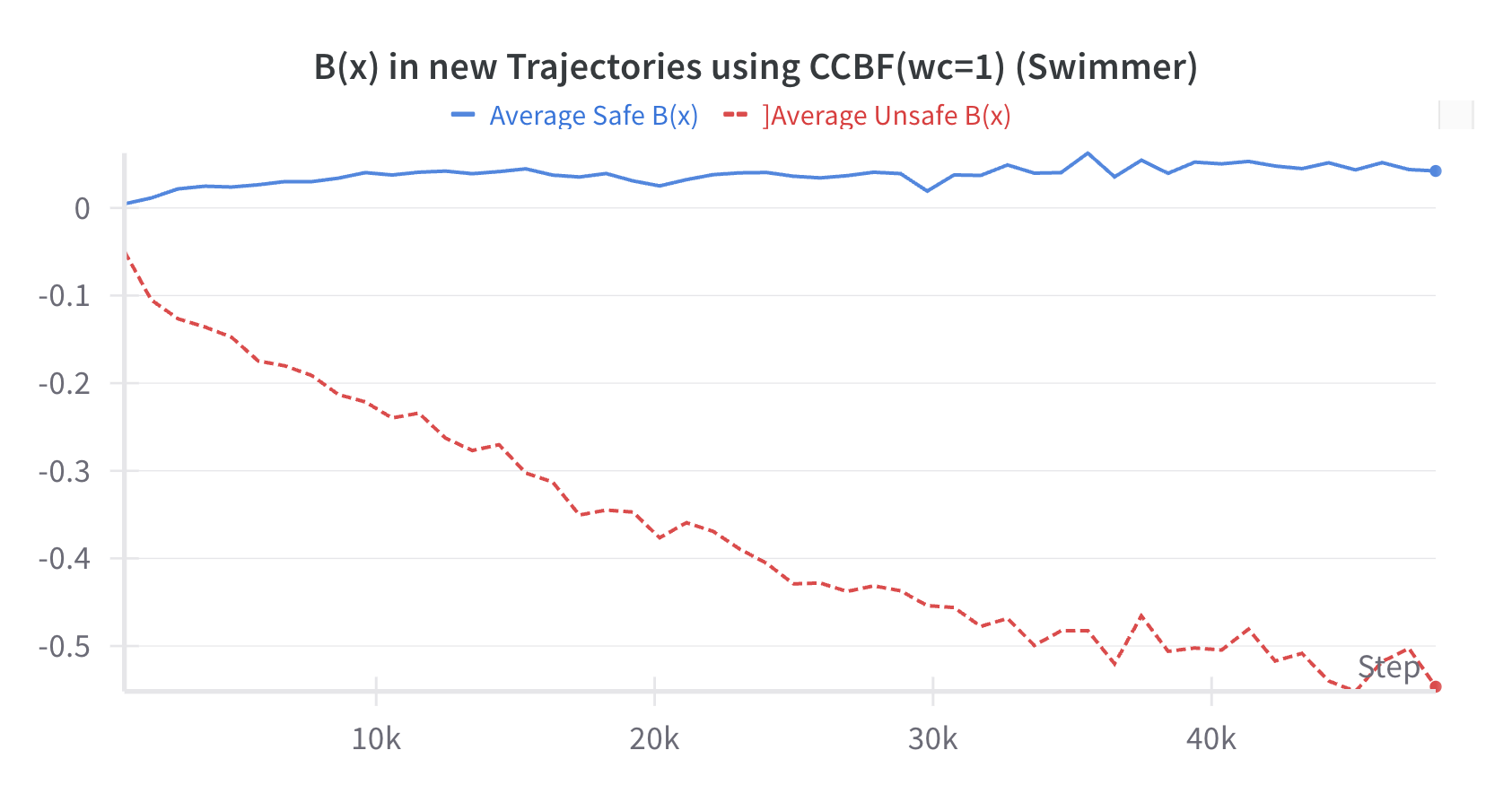} &
      \includegraphics[width=0.48\textwidth]{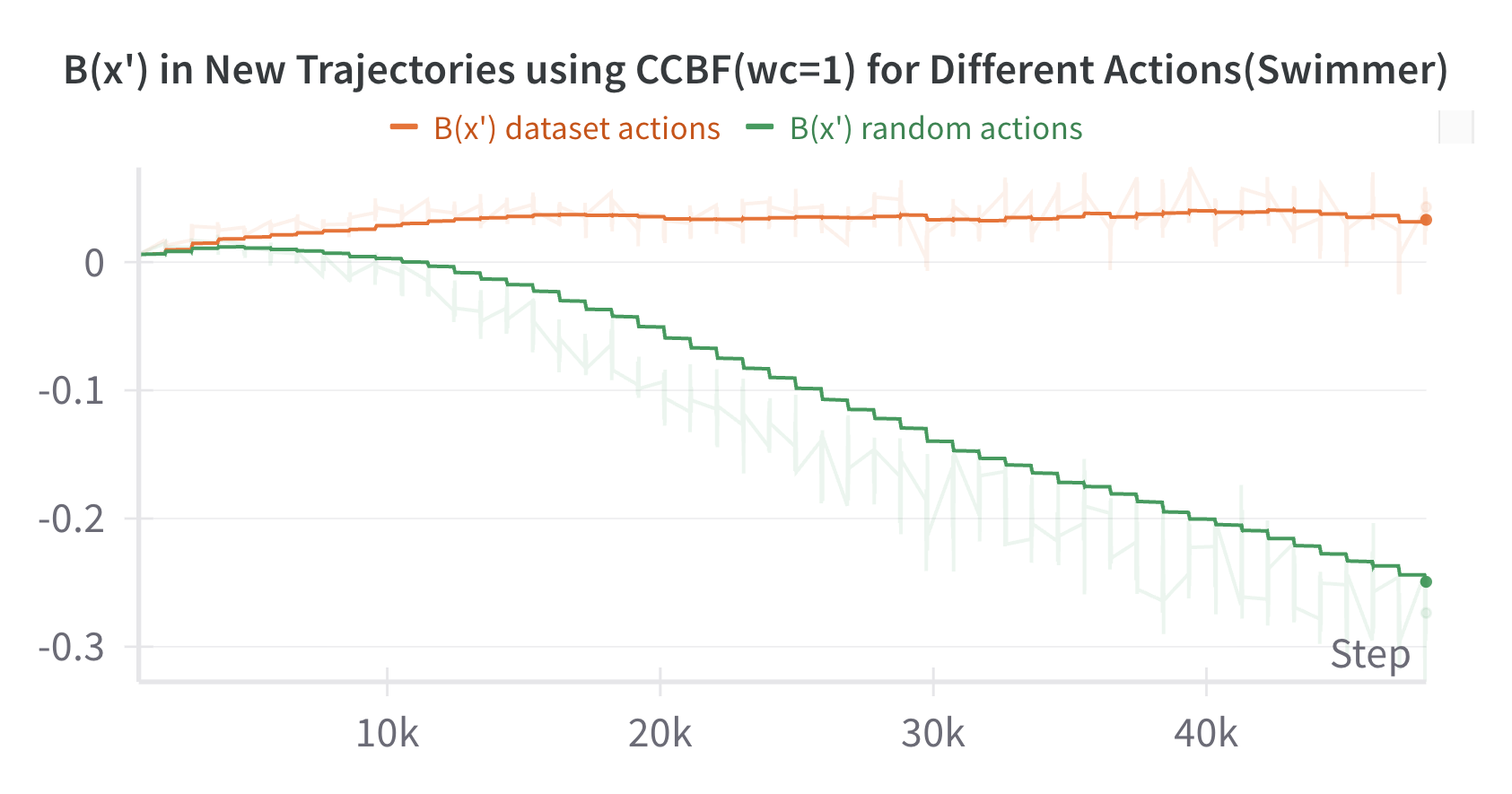} \\

      \includegraphics[width=0.48\textwidth]{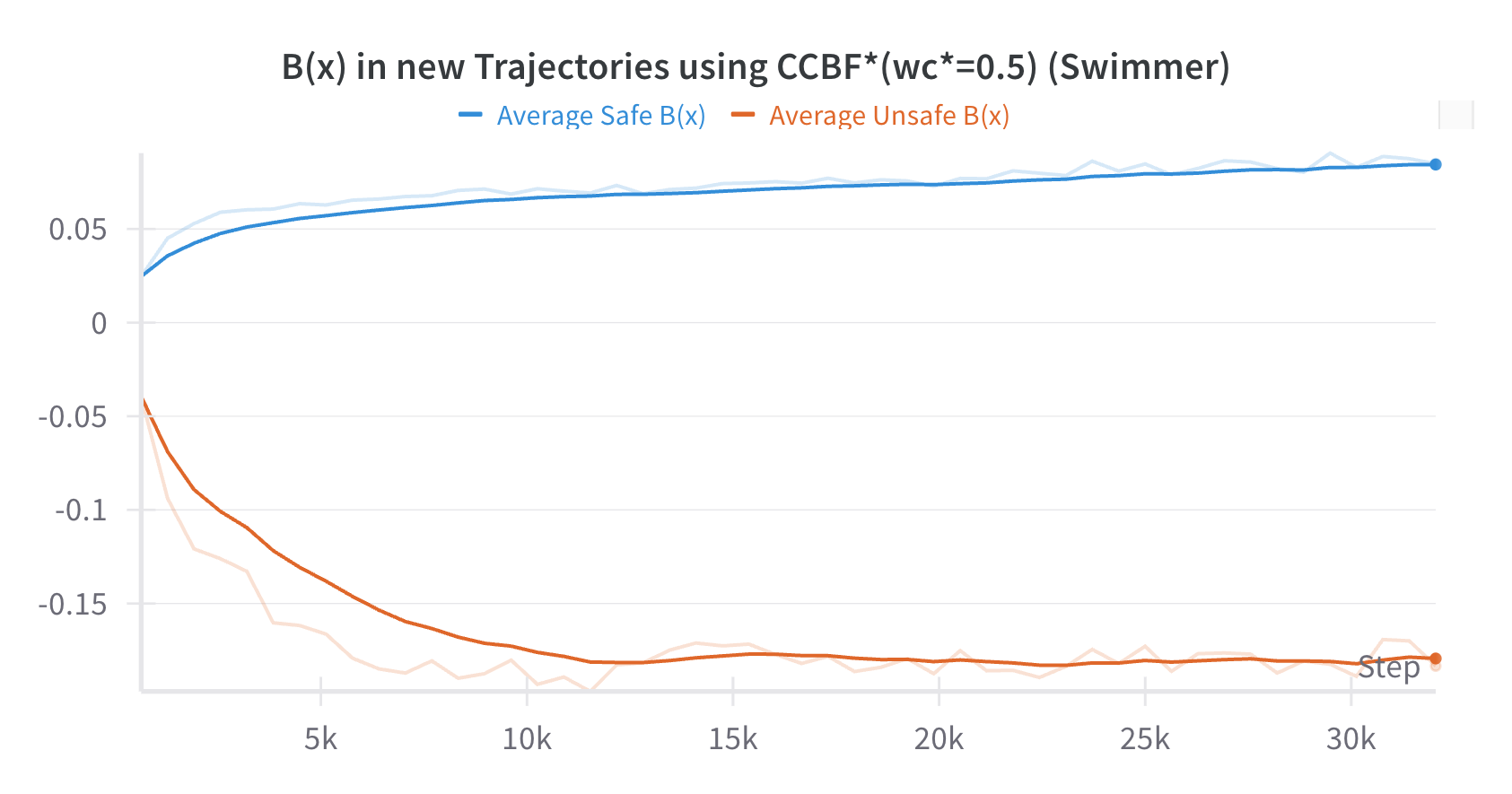} &
      \includegraphics[width=0.48\textwidth]{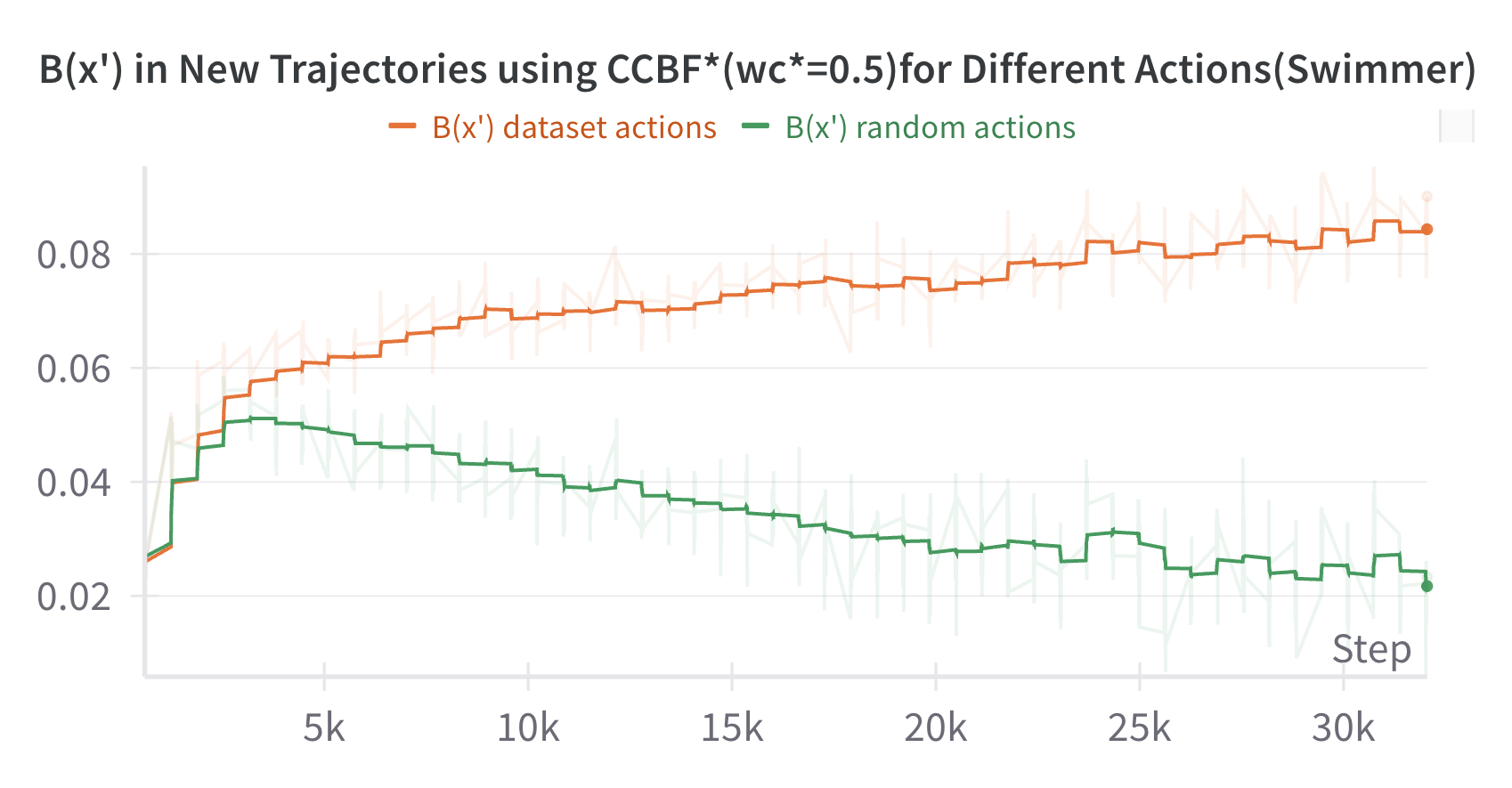} \\

      \includegraphics[width=0.48\textwidth]{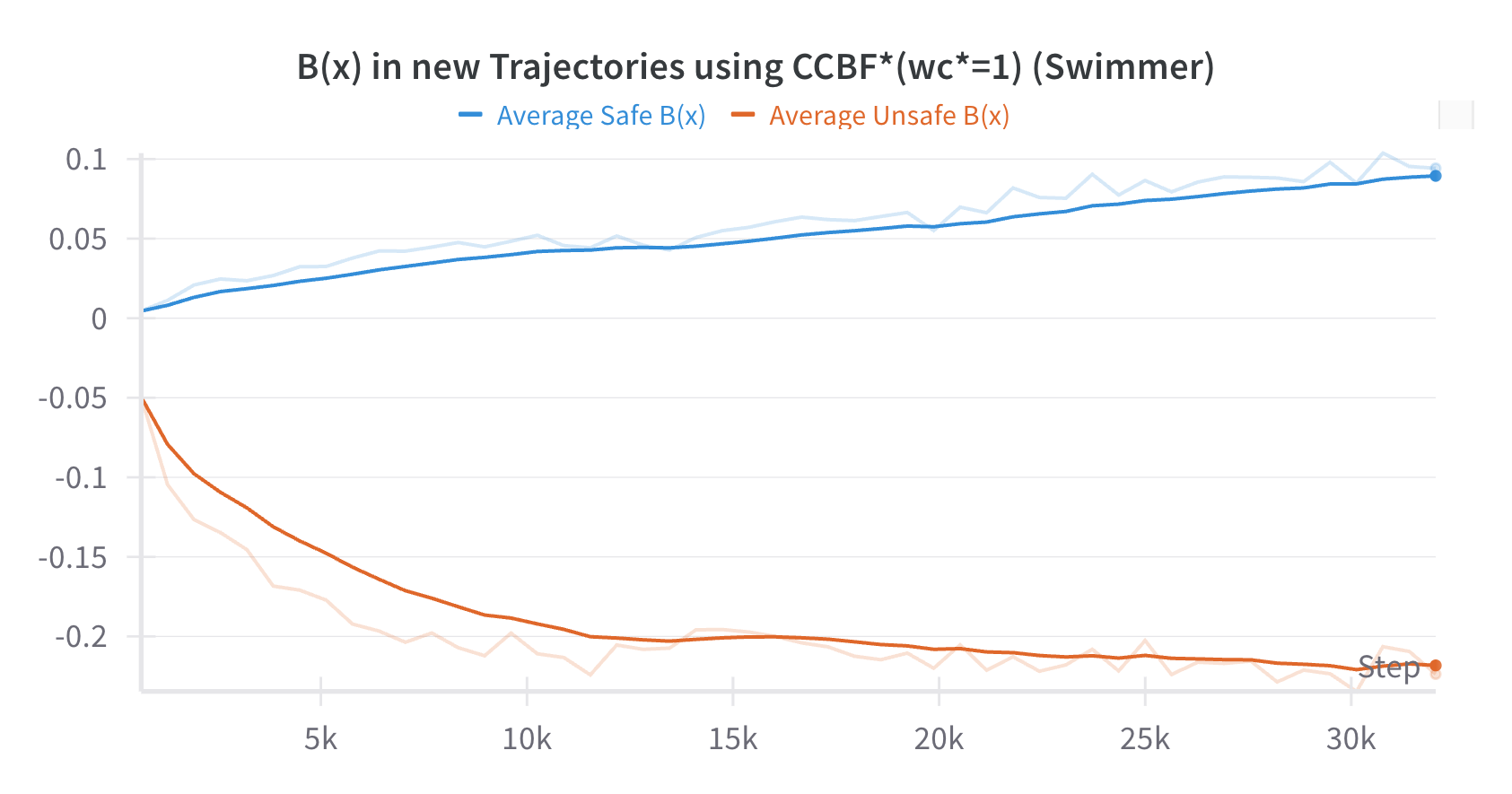} &
      \includegraphics[width=0.48\textwidth]{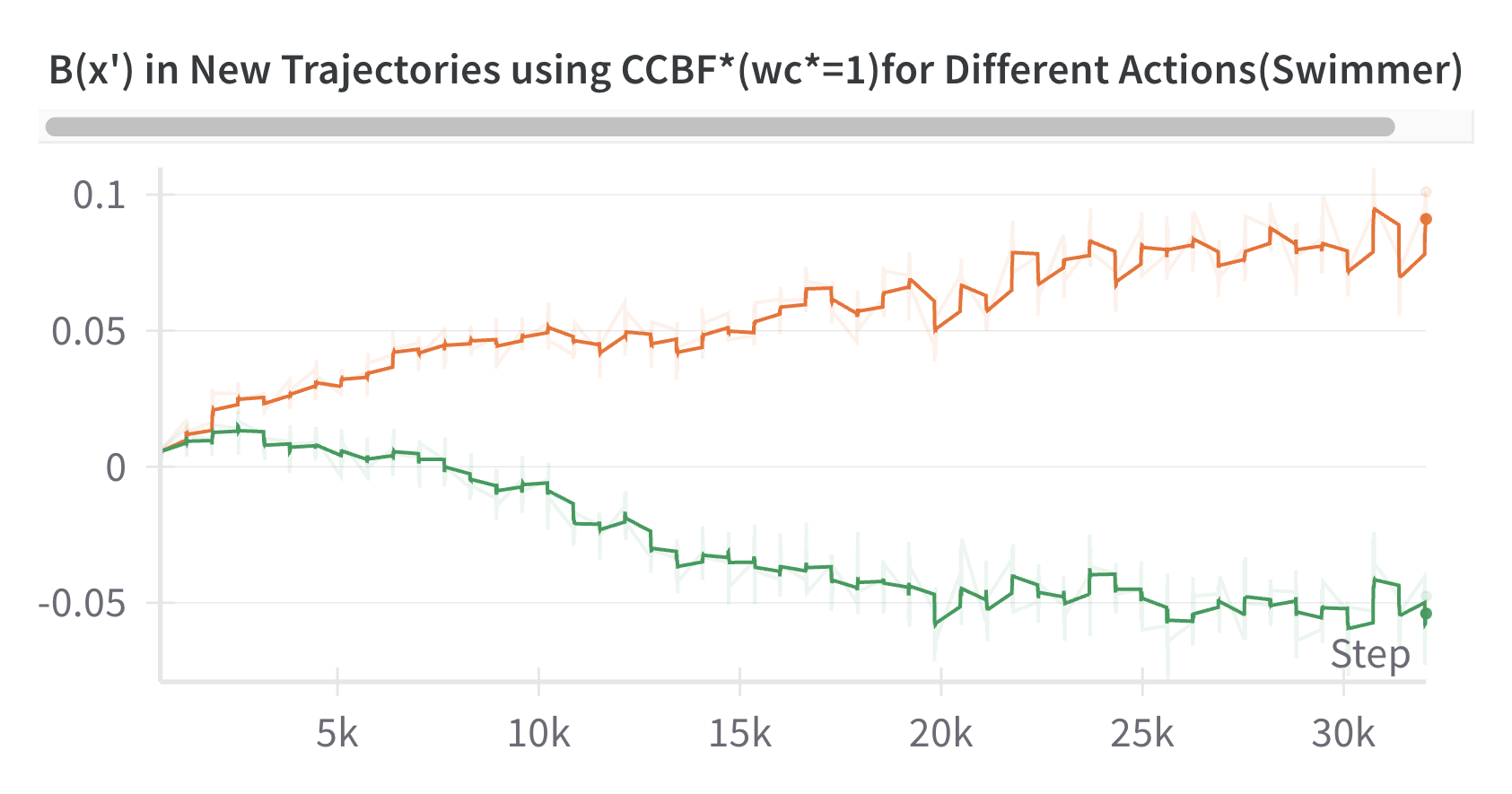} \\
    \end{tabular}
  }
  \captionof{figure}{Variation of the mean neural control barrier function values for NCBF, CCBF, and CCBF* on Swimmer test trajectories. 
  \textbf{Left column:} safe vs. unsafe states in test trajectories. 
  \textbf{Right column:} states reached by taking either dataset actions or randomly sampled actions from safe states in unseen test trajectories.}
  \label{fig:swimmer-barrier-comparison}
}]

\clearpage
\twocolumn[{%
  \centering
  \section{Comparison of the learned neural CBFs using different methods in the vision-based navigation task}
    \label{app:graphs_vision}
  \vspace{1em}
  \resizebox{\textwidth}{!}{%
    \begin{tabular}{cc}
      \includegraphics[width=0.48\textwidth]{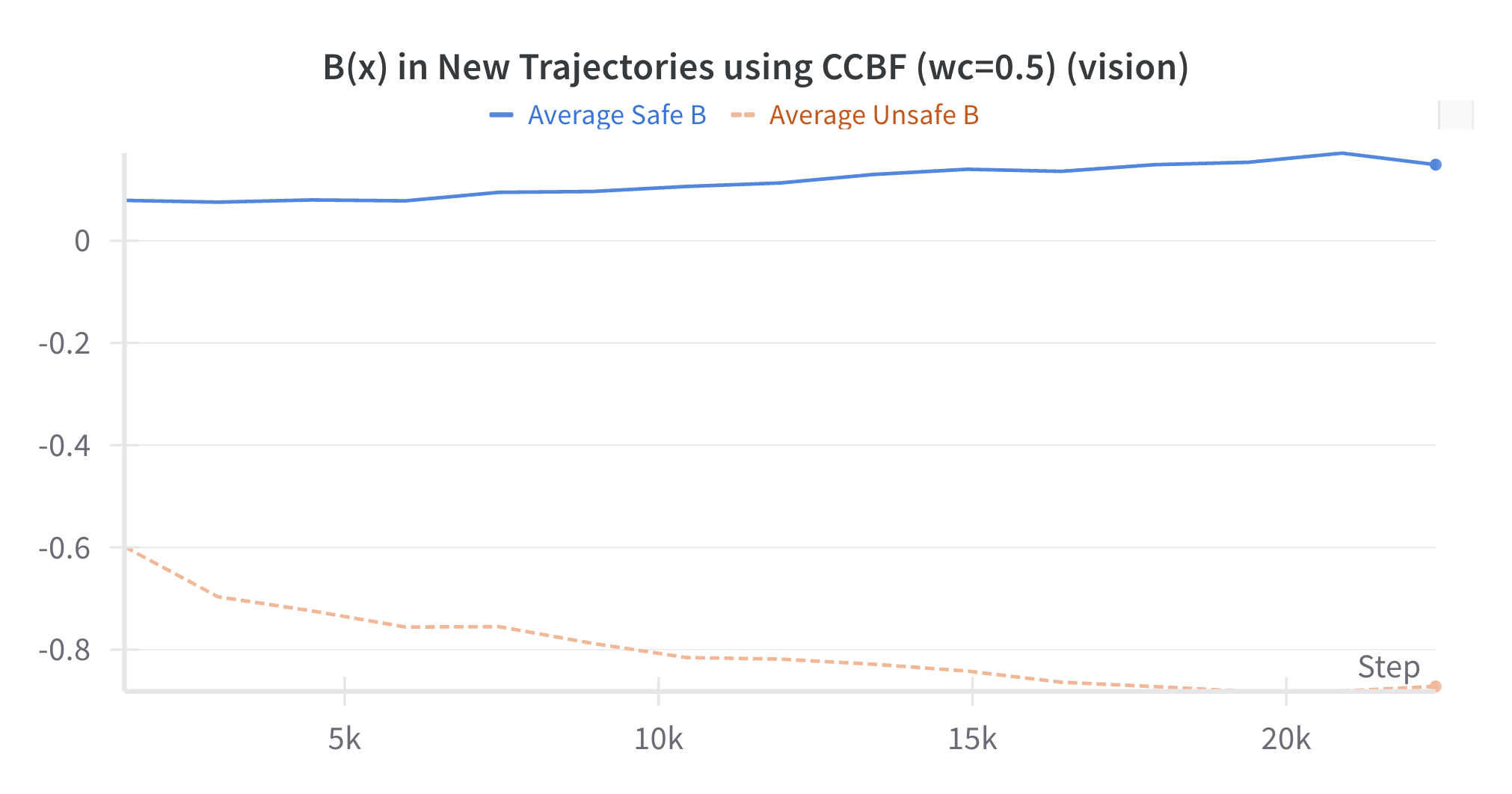} &
      \includegraphics[width=0.48\textwidth]{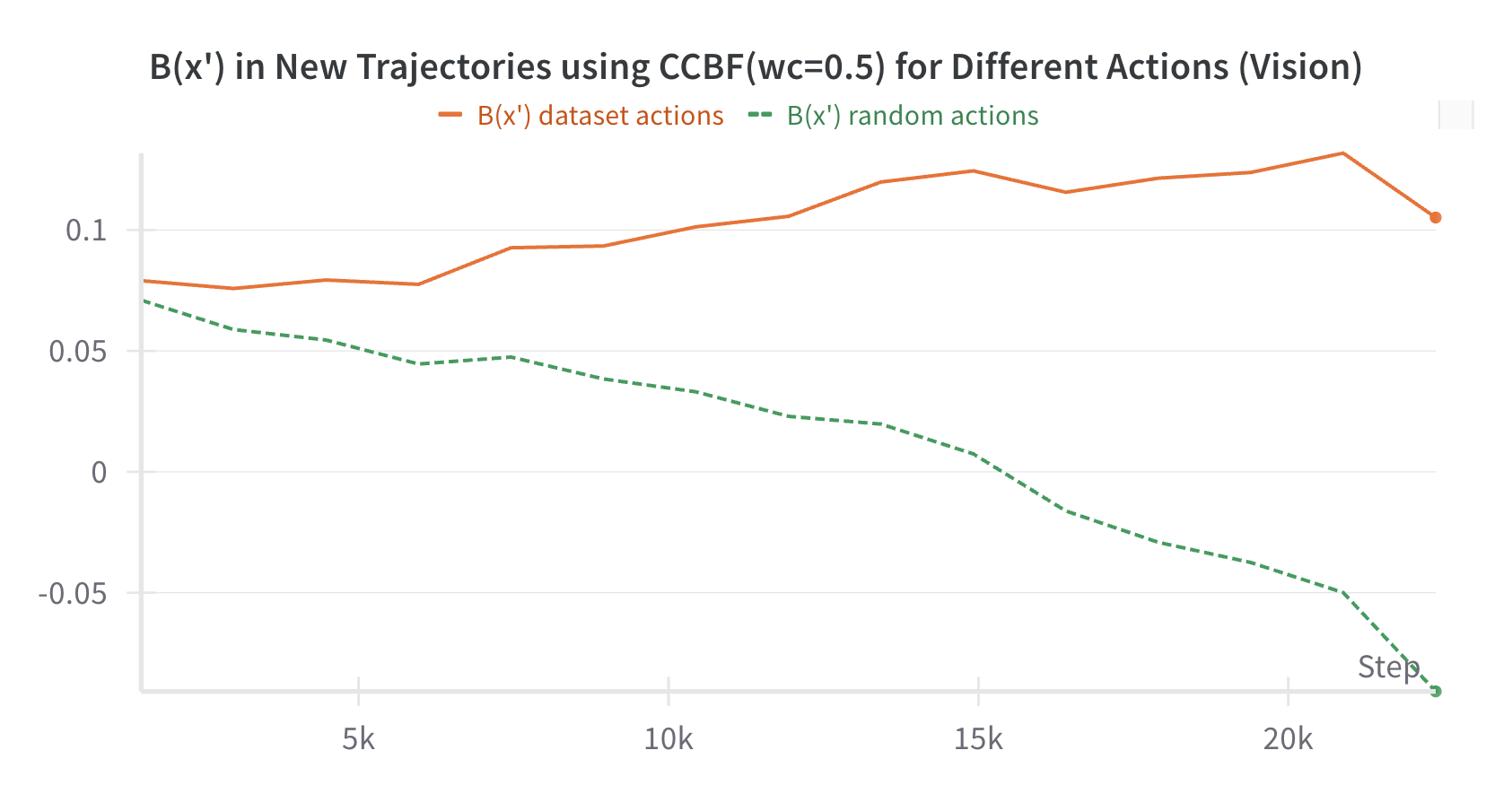} \\

      \includegraphics[width=0.48\textwidth]{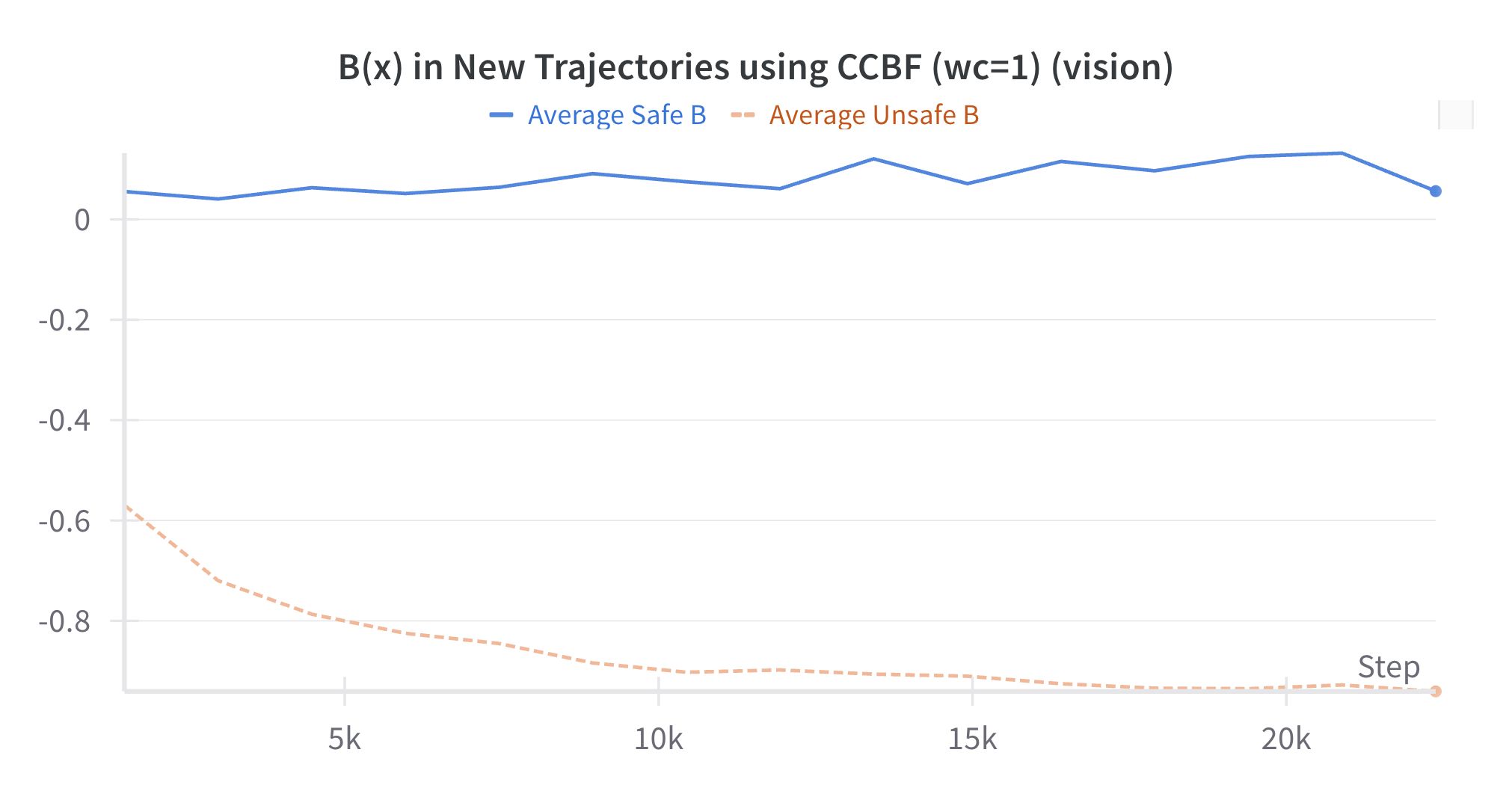} &
      \includegraphics[width=0.48\textwidth]{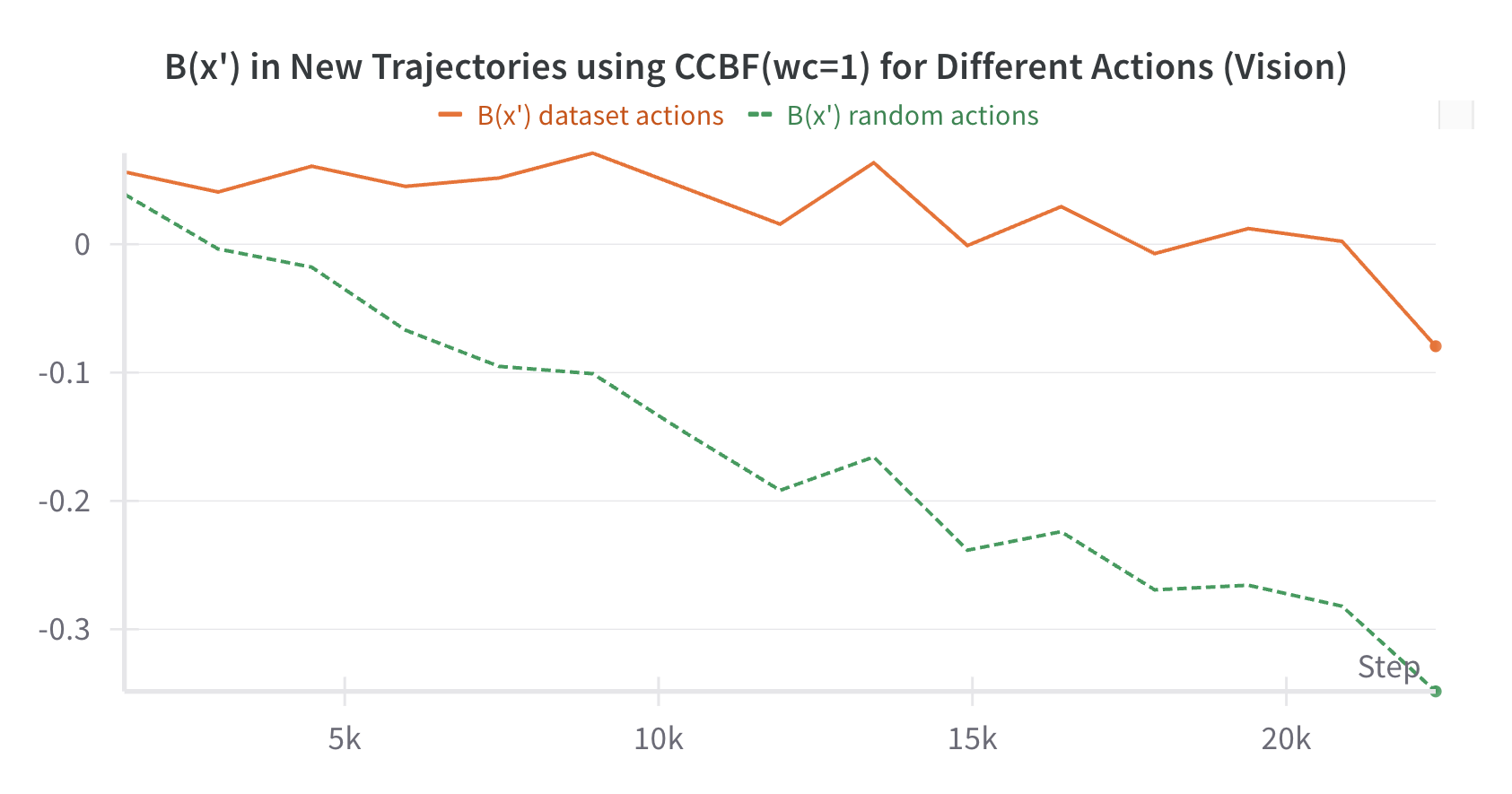} \\

      \includegraphics[width=0.48\textwidth]{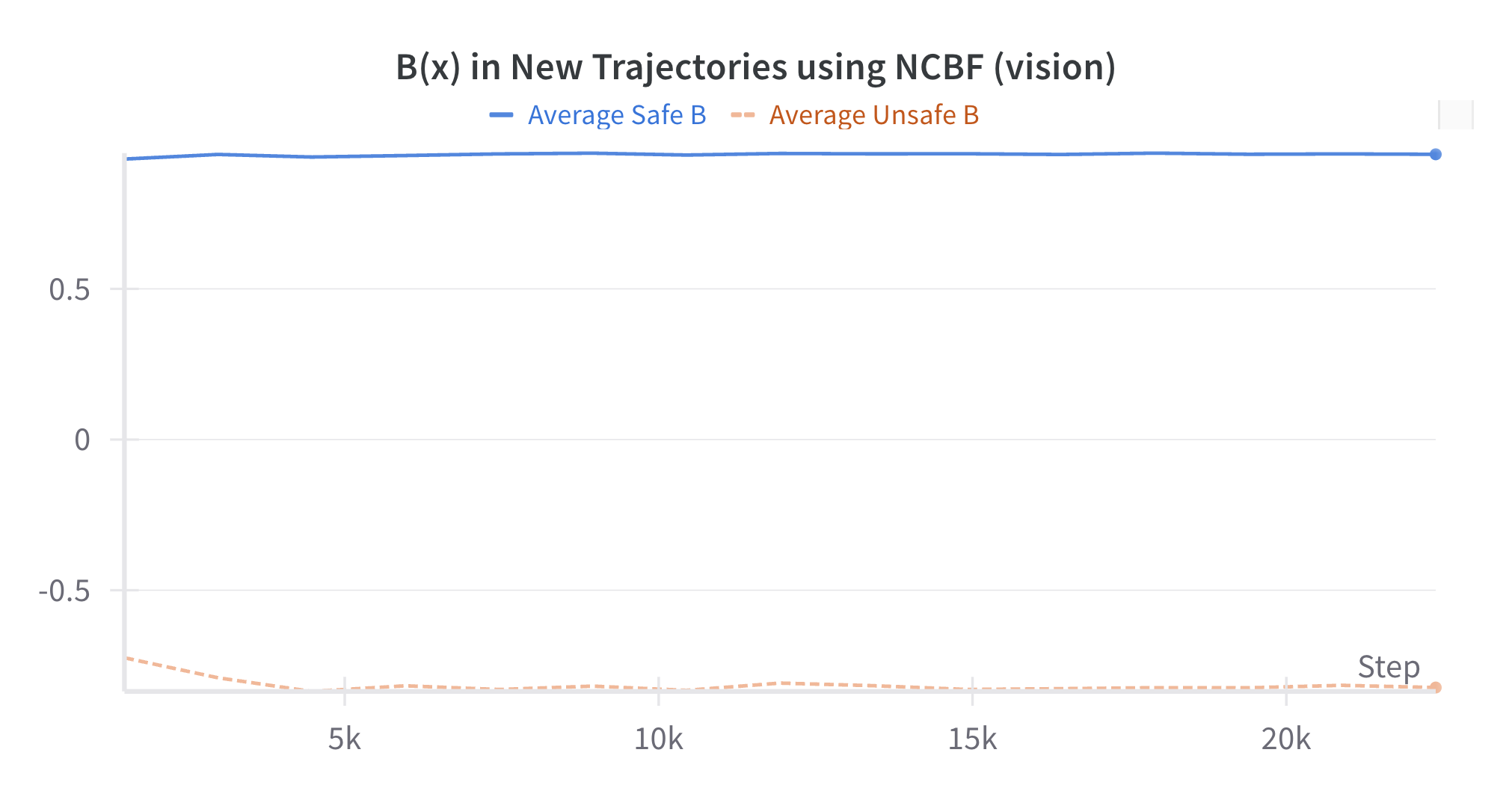} &
      \includegraphics[width=0.48\textwidth]{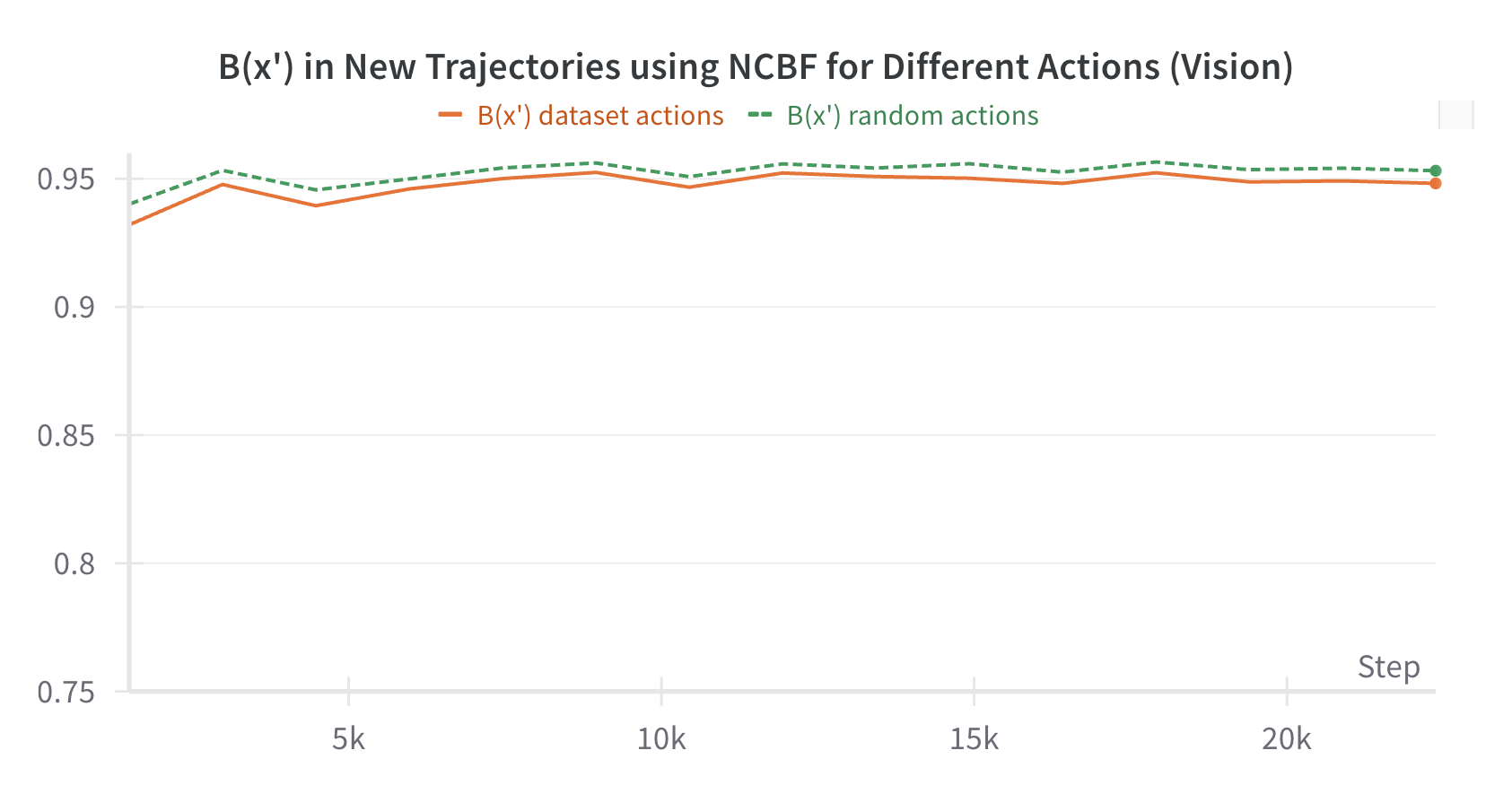} \\
    \end{tabular}
  }
  \captionof{figure}{Variation of the mean neural control barrier function values for NCBF and CCBF on vision-based navigation test trajectories. 
  \textbf{Left column:} safe vs. unsafe states in test trajectories. 
  \textbf{Right column:} states reached by taking either dataset actions or randomly sampled actions from safe states in unseen test trajectories.}
  \label{fig:vision-barrier-comparison}
}]

\end{document}